\documentclass[10pt,twocolumn,letterpaper]{article}

\usepackage{iccv}              
\usepackage{multirow}
\usepackage{colortbl}  
\usepackage{arydshln}
\usepackage{url}
\usepackage{float}
\usepackage{makecell}
\usepackage{circledsteps}
\usepackage{pifont}

\definecolor{iccvblue}{rgb}{0.21,0.49,0.74}
\usepackage[pagebackref,breaklinks,colorlinks,allcolors=iccvblue]{hyperref}

\def\paperID{3872} 
\def\confName{ICCV}
\def\confYear{2025}

\title{F-Bench: Rethinking Human Preference Evaluation Metrics for Benchmarking Face Generation, Customization, and Restoration}

\author{ Lu Liu$^{1,*}$, Huiyu Duan$^{1,*}$,  Qiang Hu$^{1,\dagger}$, Liu Yang$^{1}$,Chunlei Cai$^{2}$, \\Tianxiao Ye$^{2}$, Huayu Liu$^{1}$, Xiaoyun Zhang$^{1 }$, Guangtao Zhai$^{1}$\\
$^{1}$Shanghai Jiao Tong University, Shanghai, China $^{2}$Bilibili Inc.,Shanghai, China}

\renewcommand{\paragraph}[1]{\vspace{0pt}\noindent\textbf{#1}}
\begin{document}
\twocolumn[{%
\renewcommand\twocolumn[1][]{#1}%
\maketitle
\begin{center}
\vspace{-20pt}
    \centering
    \includegraphics[width=1\textwidth]{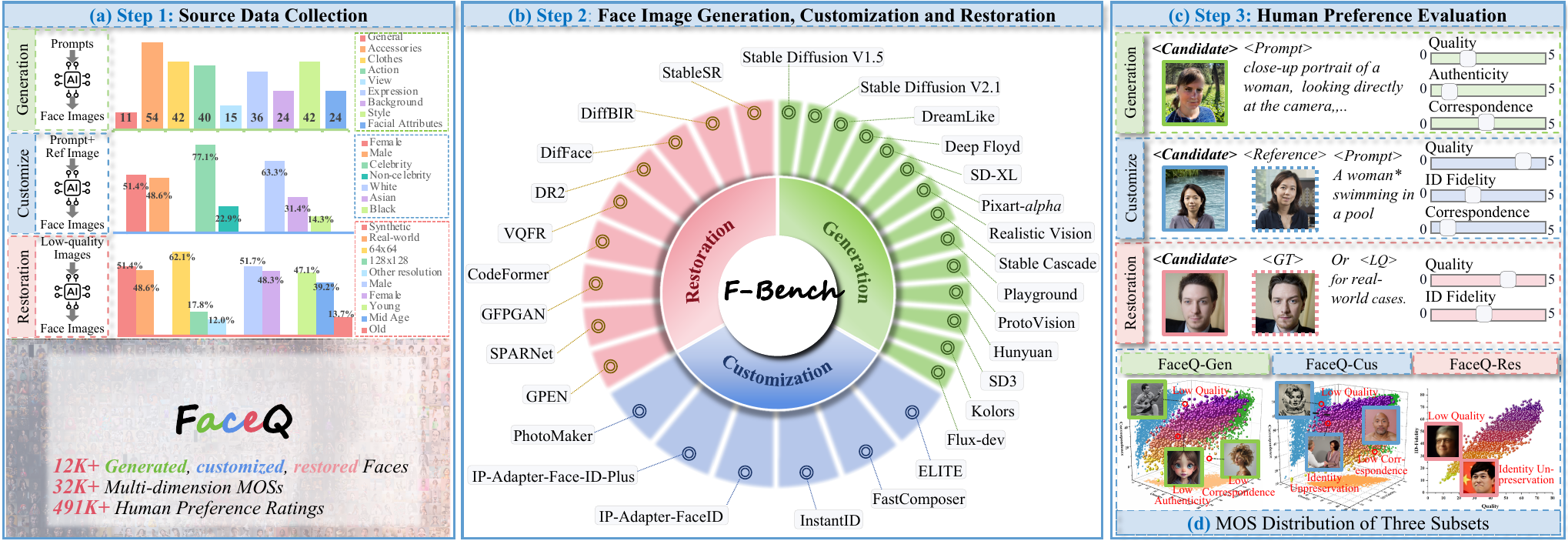}
    \vspace{-20pt}
\captionof{figure}{\textbf{Data construction pipeline of \textit{F-bench} and content overview of \textit{FaceQ} database}. From \textit{left} to \textit{right}, \textbf{(a)} Diverse input source data for face generation, customization, and restoration. \textbf{(b)} \textit{\textbf{F-Bench}} benchmarks 29 face generative models, including 14 face generation models, 6 face customization models, and 9 face restoration models. \textbf{(c)} Multi-dimensional subjective preference evaluation in \textit{\textbf{F-Bench}}. \textbf{(d)} 3D scatter plots of MOS distributions across FaceQ three subsets.}
\label{teaser}
\vspace{-3pt}
\end{center}%
}]


\maketitle
{
\renewcommand{\thefootnote}{}
\footnotetext{$^{*}$ Equal contribution. $^{\dagger}$ Corresponding author.}
}

\begin{abstract}

Recent artificial intelligence (AI) generative models have demonstrated remarkable capabilities in image production, and have been widely applied to face image generation, customization, and restoration.
However, many AI-generated faces (AIGFs) still suffer from issues such as unique distortions, unrealistic details, and unexpected identity shifts, underscoring the need for a comprehensive quality evaluation method for AIGFs.
To this end, we introduce \textbf{FaceQ}, the first comprehensive AI-generated \textbf{\underline{Face}} image database with fine-grained \textbf{\underline{Q}}uality annotations aligned with human preferences, which consists of 12K images and 491K ratings across multiple dimensions.
Using the FaceQ database, we establish \textbf{F-Bench}, a benchmark for comparing and evaluating face generation, customization, and restoration models, highlighting strengths and weaknesses across various prompts and evaluation dimensions.
Additionally, we assess the performance of existing image quality assessment (IQA) methods on FaceQ, and further propose a large multimodal model (LMM) based \textbf{\underline{F}}ace quality \textbf{\underline{Eval}}uator (\textbf{F-Eval}) to accurately assess the multi-dimensional quality of generated faces in a one-for-all manner.
Extensive experimental results demonstrate the state-of-the-art performance of our F-Eval.
The project page is: \url{https://mediax-sjtu.github.io/F-Bench/}.

\end{abstract}

\section{Introduction}

\label{sec:intro}
\begin{table*}[!t]
  \centering
  \caption{Comparison between \textbf{FaceQ} and existing AIGC quality assessment databases \textit{(top)} as well as natural face quality assessment databases \textit{(bottom)}. FaceQ is the first large-scale AI-generated, customized and restored face quality assessment database.}
  \vspace{-10pt}
  \renewcommand\arraystretch{1}
  \resizebox{\textwidth}{!}{
    \begin{tabular}{lccccccc}
      \toprule[1pt]
      \bf Database &Domain& \bf Source  &  \bf Images& \bf Scores & \bf  Dimensions \\
      \midrule
    AGIQA-3K (\textit{TCSVT2024})~\cite{li2023agiqa3kopendatabaseaigenerateliang2024richd}&General&AI-generated&2,982&MOS& Perception, Alignment\\
    AIGCIQA (\textit{CVPR2024})~\cite{wang2023aigciqa2023largescaleimagequality}&General&AI-generated&2,400&MOS& Quality, Authenticity, Correspondence\\
      RichHF-18K (\textit{CVPR2024})~\cite{liang2024rich}&General&AI-generated&17,760&MOS&Plausibility, Aesthetics, Text-image Alignment, Overall\\
      \hdashline
      PIQ23 (\textit{CVPR2023})~\cite{chahine2023imagequalityassessmentdataset} &Face& Real  & 5,116 & Pair& Overall, Details, Exposure  \\
      \rowcolor{gray!20} \bf FaceQ &\bf Face& \bf AI-generated, customized and restored  & \bf 12,255 & \bf MOS & \bf Quality, Authenticity, ID Fidelity, Correspondence  \\
      \bottomrule[1pt]
    \end{tabular}
    }
\vspace{-18pt}
  \label{database}
\end{table*}

Generative models, including VAE~\cite{vae}, GAN~\cite{gan}, and diffusion models~\cite{ho2020denoising,song2020denoising,Rombach_2022_CVPR}, have made remarkable progress in recent years, and have been extended to the fields of text-to-image generation \cite{flux}, text-driven image editing \cite{kawar2023imagic}, image restoration \cite{xia2023diffir}, \textit{etc}, leading to high-quality artificial intelligence generated images (AIGIs).
As an important aspect of AIGIs, many studies have also investigated face generation \cite{xia2021tedigan}, face customization~\cite{li2023photomaker,ye2023ip}, and face restoration~\cite{lin2023diffbir,wang2023exploiting}, producing creative, high-quality face images. 
Despite these advances, many AI-generated faces (AIGFs) still suffer from perceptual deficiencies.

Many natural face image quality assessment databases have been established in the literature~\cite{kirstain2023pickapicopendatasetuser,li2023agiqa3kopendatabaseaigenerateliang2024richd,wang2023aigciqa2023largescaleimagequality,liang2024rich}.
However, natural images are mainly degraded by distortions such as noise, blur, compression \cite{zhai2020perceptual,min2024perceptual,duan2024finevq}, \textit{etc.}, while AIGIs generally suffer from unique distortions including unrealistic structures and text-image inconsistencies \cite{liang2024rich,wang2023aigciqa2023largescaleimagequality}, \textit{etc}.
Recently, many studies have investigated human preferences for AIGIs \cite{liang2024rich,wang2023aigciqa2023largescaleimagequality,li2023agiqa,huang2023t2i,imagereward}.
However, these general AIGI quality assessment works may not be well aligned with human preferences on AIGFs, since the perceptual quality of AIGFs can be significantly affected by small deviations in facial attributes,(\textit{e.g.}, eye size, skin texture, identity shifts), while the number of AIGFs in these AIGI databases is limited.
Moreover, current AIGI quality assessment works mainly focus on the image generation, while human preferences for face customization have rarely been considered, where identity fidelity is significant.

The above limitation highlights the necessity of establishing a comprehensive quality evaluator for AIGFs.
To this end, we propose \textbf{FaceQ}, the first AI-generated \textbf{\underline{Face}} image \textbf{\underline{Q}}uality assessment database with three popular tasks. Specifically, FaceQ contains 12K+ images obtained from 29 generative models across three tasks: 1) face generation, 2) face customization, and 3) face restoration, as shown in Fig.\ref{teaser}. Based on the generated images, we collect 491K+ quality ratings from multiple dimensions including quality, authenticity, identity (ID) fidelity, and text-image correspondence, respectively, finally forming 32K+ mean opinion scores (MOSs). 
Using FaceQ, we establish a human preference benchmark, \textit{i.e.}, \textbf{F-Bench}, for face generation, customization, and restoration models, and thoroughly analyze the generation performance across different evaluation perspectives, models, and categories.
In addition, we also compare the correlation of existing image quality assessment (IQA), face quality assessment (FQA), AIGI quality assessment, and preference evaluation metrics with human preferences.

Furthermore, we present \textbf{F-Eval}, the first generated \textbf{\underline{F}}ace image quality \textbf{\underline{Eval}}uation method designed to perform comprehensive multi-dimensional quality evaluation for generated, customized, and restorted faces in a \textbf{one-for-all} manner. 
Specifically, F-Eval leverages large multimodal models (LMMs) \cite{bai2025qwen2} and adopts instruction tuning~\cite{liu2023visual} for preference-related vision-language adaption.
We also apply low-rank adaptation (LoRA)~\cite{hulora} techniques for more efficient feature refining.
Extensive experimental results demonstrate that F-Eval outperforms existing quality assessment methods in terms of multiple dimensions relevant to human preference for all three tasks. The main contributions of our work are highlighted as follows:

\begin{itemize}
    \item We construct \textbf{FaceQ}, the first generated face quality assessment database consisting 491K+ human preference ratings from quality, authenticity, ID fidelity, correspondence dimensions, on 12K+ generative face images across three tasks.
    \item We establish \textbf{F-Bench}, a comprehensive human preference benchmark for 29 face generation, customization, and restoration methods. It also extensively evaluates the alignment of 26 existing quality assessment methods with human preferences.
    \item We present \textbf{F-Eval}, a LMM-based quality assessment model to predict multi-dimensional human preference on generated faces. Extensive experiments demonstrate the state-of-art QA performance of the proposed F-Eval.
\end{itemize}

\section{Related Work}
\label{sec:rw}
\begin{figure*}
    \centering
    \includegraphics[width=1\linewidth]{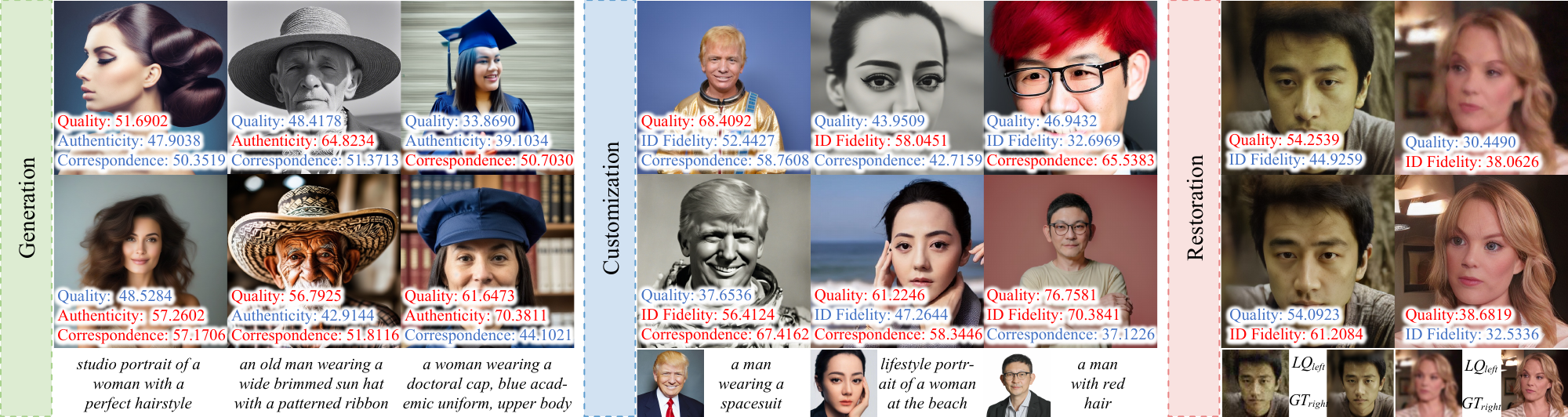}
    \vspace{-21pt}
    \caption{\textbf{Rating comparisons of eight dimensions.} Each column presents a pair of intuitive examples of each dimension, with  \textcolor{red}{red} indicating the better rating and \textcolor{blue}{blue} indicating the worse one. From \textit{left} to \textit{right}, the subsets are face generation, face customization, and face restoration subsets. The last row displays the corresponding prompts, reference image-prompt pairs, and the GT-LQ image pairs. }
    \label{fig:ratings}
    \vspace{-17pt}
\end{figure*}
\paragraph{Face Generation \& Customization \& Restoration.} \textbf{(1) }For face generation models, earlier works such as GANs~\cite{gan, karras2019style,viazovetskyi2020stylegan2} have demonstrated significant influences. Recently, diffusion models~\cite{Rombach_2022_CVPR,saharia2022photorealistic,ramesh2022h} have achieved rapid developments in text-to-image generation. These models can be categorized into pixel-space diffusion models ~\cite{ramesh2021zeroshottexttoimagegeneration,ramesh2022h,saharia2022photorealistictexttoimagediffusionmodels, IF}, and latent space diffusion models~\cite{Rombach_2022_CVPR,li2024hunyuanditpowerfulmultiresolutiondiffusion,podell2023sdxlimprovinglatentdiffusion,chen2023pixartalphafasttrainingdiffusion}. \textbf{(2) }For face customization, this is a recently defined task that aims to create images of a specific person with the input of an identity reference image and prompts~\cite{wei2023eliteencodingvisualconcepts}. It can be divided into two categories including conventional test-time optimization~\cite{ruiz2023dreambooth,gal2022image,kumari2023multi} and tuning-free customization methods utilizing pre-trained diffusion models~\cite{xiao2023fastcomposertuningfreemultisubjectimage,ye2023ip,li2023photomaker,cui2024idadapter,he2024id,wang2024instantid}. 
\textbf{(3)} For face restoration, which aims to restore high-quality face images from their low-quality counterparts. Some traditional restoration methods utilize geometric priors ~\cite{chen2018fsrnet, kim2019progressive,chen2021progressive} or reference priors~\cite{Li_2018_ECCV,li2020enhanced} while in recent years, exploring generative prior such as GAN priors~\cite{wang2021gfpgan,yang2021gan,chen2020learning,gu2022vqfr,zhou2022towards} and diffusion priors~\cite{wang2023dr2,wang2023exploiting,lin2023diffbir,yue2022difface,tian2025towards,hu2025varfvv,Hu_2025_CVPR} has become a prevalent trend. Despite these improvements, the generated, customized, and restored images may suffer from various quality issues, highlighting the necessity of evaluators.\\
\paragraph{Face Image Quality Assessment Database.}
Existing face image quality assessment (FIQA) can be divided into Biometric FIQA~\cite{boutros2023cr,hernandez2019faceqnet,lijun2019multi,ou2021sdd} and Generic FIQA~\cite{wang2022surveydeepfacerestoration,gfiqa20k,chahine2023imagequalityassessmentdataset,dsl-fiqa}.
BFIQA methods are most developed to evaluate the biometric utility for  face recognization. GFIQA methods~\cite{gfiqa20k} focus on perceptual degradation in face restoration task. Both of these databases consist of natural face images, which has distinctly different nature with generated faces. \\
\paragraph{AIGC Quality Assessment Database.}
In recent years, plenty of researchers have raised interest in assessing AI-generated images(AIGIs) quality assessment ~\cite{wang2023aigciqa2023largescaleimagequality,kirstain2023pickapicopendatasetuser,li2023agiqa3kopendatabaseaigenerateliang2024richd,duan2024finevq,yang2024aigcoiqa2024,jiangcatekv} by building various databases as shown in Tab~\ref{database}. All of these existing database contain various categories of general AIGI images, leaving a notable gap in datasets dedicated to AI-generated face images.

\section{\textit{FaceQ} Database Construction}
\label{sec:methods}
In this section, we introduce the proposed generative face quality assessment database \textbf{FaceQ}. The database includes 12,255 images and 491,130 human opinion ratings. Our FaceQ demonstrates the merits of comprehensive generated visual content and multi-dimensional labeling, which can significantly facilitate the advancement of AIGC research.

\begin{figure*}[h]
\end{figure*}
\begin{figure*}[h]
    \centering
    \includegraphics[width=1\linewidth]{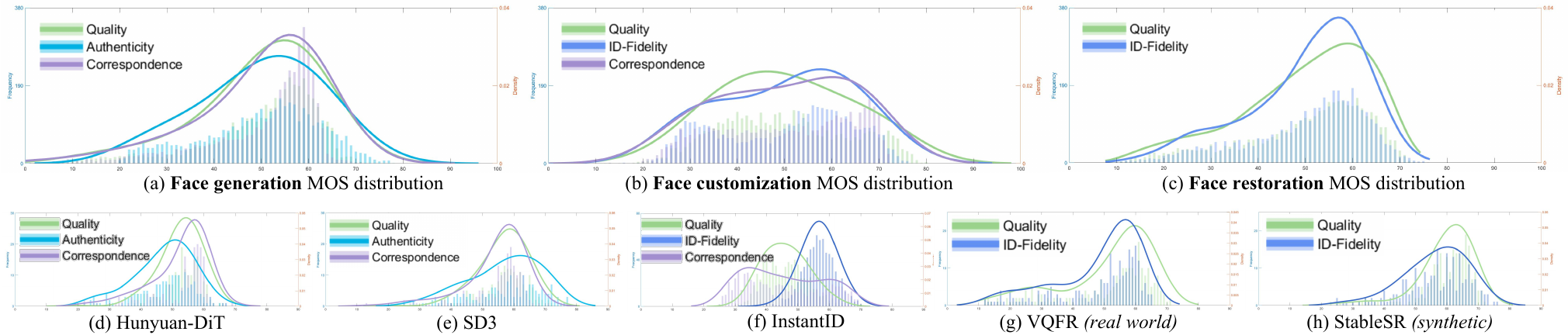}
        \vspace{-20pt}
    \caption{\textbf{Mean Opinion Score (MOS) distribution histograms and kernel density curves.} (a)-(c) MOS distributions for three subsets. (d)-(h) Model-wise MOS distributions. MOS Distributions of other methods are provided in the \textit{supplementary material}.}
    \label{fig:mos}
    \vspace{-10pt}
\end{figure*} 
\subsection{Data Collection}
\subsubsection{\textbf{\textit{FaceQ-Gen}} Subset Collection }
\label{FaceQ-Gen}
\paragraph{Prompt Sources.} 
The prompts in \textit{FaceQ-Gen} subset are sourced from MS-COCO~\cite{DBLP:journals/corr/LinMBHPRDZ14} 
~\cite{li2023photomaker} and GPT-4o~\cite{gpt}. These face-centric prompts can be categorized into nine types: \textit{General, Accessories, Clothes, Action, View, Expression, Background, Style}, and \textit{Facial Attributes}. \\
\paragraph{Models Collection.} We collect 14  representative face generation models, including Deep Floyd~\cite{IF}, Stable Diffusion V1.5~\cite{Rombach_2022_CVPR}, Stable Diffusion V2.1~\cite{Rombach_2022_CVPR}, SD-XL~\cite{podell2023sdxlimprovinglatentdiffusion}, Realistic Vision~\cite{realistivision}, Stable Cascade~\cite{stablecascade}, ProtoVision V6.6~\cite{protovision}, SD3~\cite{sd3}, Playground V2.5~\cite{li2024playgroundv25insightsenhancing}, Kolors~\cite{kolors}, PixArt-alpha~\cite{chen2023pixartalphafasttrainingdiffusion}, Hunyuan~\cite{li2024hunyuanditpowerfulmultiresolutiondiffusion}, and Flux-dev~\cite{flux}. Negative prompts and safe sensors are enabled. In total, \textit{FaceQ-Gen} includes 4,032 generated face images. 

\subsubsection{\textbf{\textit{FaceQ-Cus} Subset Collection}}
\paragraph{Identity and Prompt Sources.}
Face customization relies on a reference identity image and a guiding prompt. The reference images in \textit{FaceQ-Cus} are partially self-collected, with the rest sourced from~\cite{li2023photomaker}. The prompts are selected and filtered from the \textit{FaceQ-Gen} prompt list.

\paragraph{Models Collection.}
We collect six representative face customization methods, including ELITE~\cite{wei2023eliteencodingvisualconcepts}, FastComposer~\cite{xiao2023fastcomposertuningfreemultisubjectimage}, IP-Adapter-FaceID~\cite{ipa1}, InstantID~\cite{wang2024instantid}, IP-Adapter-Face-ID-Plus~\cite{ipa2}, and PhotoMaker V2~\cite{li2023photomaker}. Here, we use single reference image as input. In total, \textit{FaceQ-Cus} includes 4,200 generated face images. 

\subsubsection{\textbf{\textit{FaceQ-Res}} Subset Collection}
\paragraph{Low Quality Images Source.}
\textbf{(1)} \textbf{Synthetic}: We construct two degraded pipelines following previous works~\cite{chen2020learning,wang2021towards,wang2023dr2,wang2021realesrgan} to simulate real-world degradations on both in-domain data from VFHQ~\cite{wang2022vfhqhighqualitydatasetbenchmark}, CelebRef~\cite{li2022learning} and out-of-domain self-collected high-quality Asian face images  \textbf{(2) }\textbf{Real-world}: Real-world low-quality images are sourced from various public datasets, including AgeAB~\cite{moschoglou2017agedb}, CelebChild~\cite{gu2022vqfr}, Wider~\cite{yang2016wider}, LFW~\cite{LFWTech} and MegaFace~\cite{kemelmacher2016megaface}, ensuring the degradation diversity. 

\paragraph{Models Collection and Face Image Restoration.}
We evaluate both diffusion-prior-based methods such as DiffBIR~\cite{lin2023diffbir}, DifFace~\cite{yue2022difface}, StableSR~\cite{wang2023exploiting}, and GAN or Transformer-based methods GFPGAN~\cite{wang2021gfpgan}, GPEN~\cite{yang2021gan}, CodeFormer~\cite{zhou2022towards}, VQFR~\cite{gu2022vqfr} and Face-SPARNet~\cite{chen2020learning}. Each method is benchmarked on both the \textit{F-Res} synthetic dataset and the real-world dataset. In total,\textit{FaceQ-Res} include \textbf{4023} restored images, with 2007 images in the real-world case and 2016 images in the synthetic case. 

\subsection{Human Preference Dimension Design}
The evaluation dimensions of generated faces need specific design to adequately reflect human preference, given the complexity of perceptual factors involved. To the end, we decompose “quality” into four core dimensions: \textit{Quality}, \textit{Authenticity}, \textit{ID Fidelity}, and \textit{Correspondence}. \\
\paragraph{\textit{\textbf{Quality}}.} This dimension evaluates the overall perceptual quality of the image, considering factors such as color accuracy, blur, noise, and artifacts. It is used in all tasks.\\
\paragraph{\textit{\textbf{Authenticity}}.} This dimension evaluates how closely the generated image resembles a natural, real-life photograph, with particular emphasis on realistic skin texture, facial details, and wrinkles. It is used in face generation task.\\
\paragraph{\textit{\textbf{ID Fidelity}}.}  This dimension evaluates how well the identity of the reference image is preserved in the customized or restored image in face customization and restoration tasks.\\
\paragraph{\textit{\textbf{Correspondence}}.} This dimension evaluates text-to-image alignment in face generation and customization tasks.

\subsection{Subjective Experiment}
\paragraph{Main Assessment.}
To ensure a comprehensive, fair, and reliable subjective assessment, 180 participants with normal or corrected-to-normal eyesight are recruited for the experiment. They first take expert-led tutorials before the main assessment. In the main assessment, participants are asked to rate the image by dragging a sliding window from 0 to 5 with a two-digit decimal from different dimensions. All images are displayed on 27-inch 4K Dell monitors, randomly presented, at their original resolution under standard lighting conditions. The experiment follows the ITU-R BT.500-14 guidelines~\cite{series2012methodology} for subjective evaluations.

\paragraph{Subjective Score Processing.}
To eliminate the inter-annotator variability, outlier detection based on Kurtosis is conducted with a rejection rate of 3\% ~\cite{series2012methodology}. As a result, each image is rated by at least 15 subjects. These valid ratings are converted and scaled into Z-scores ranging from $[0,100]$ with the following formulas: 
\begin{eqnarray}
\begin{aligned}
z_{ij}=\frac{r_{ij}-\mu_i}{\sigma_i}, ~\quad z_{ij}'=\frac{100(z_{ij}+3)}{6},
\end{aligned}
\end{eqnarray}
\begin{eqnarray}
\begin{aligned}
\mu_i=\frac{1}{M_i}\sum_{j=1}^{M_i}r_{ij}, ~\sigma_i=\sqrt{\frac{1}{M_i-1}\sum_{j=1}^{M_i}{(r_{ij}-\mu_i)^2}},
\end{aligned}
\end{eqnarray}

Here, \( r_{ij} \) represents the original ratings provided by the \( i \)-th evaluator for the  \( j \)-th image,\( M_i \) denotes the total number of images that were assessed by the \( i \)-th participant in this group.
Finally, the mean opinion score (MOS) of the image \(j\) $(MOS_j)$  is computed by averaging the rescaled z-scores as follows:
\begin{eqnarray}
\begin{aligned}
MOS_j=\frac{1}{N_j}\sum_{i=1}^{N_j}z_{ij}'
\end{aligned}
\end{eqnarray}
where $N_j$ denotes the number of valid subjects for image $j$, and $z_{ij}'$ denotes rescaled z-scores.

\begin{figure*}[t]
    \centering
    \includegraphics[width=1\linewidth]{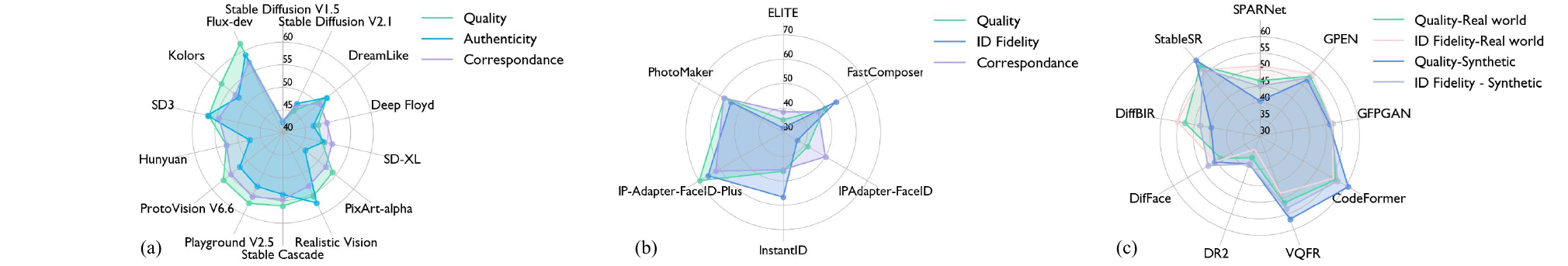}
      \vspace{-17pt}
    \caption{\textbf{Average MOS score comparison across all models and dimensions.} (a) Face generation. (b) Face customization. (c) Face restoration. The models are arranged in a clockwise order by release date.}
    \label{fig:radar}
 \vspace{-8pt}
 \end{figure*}
\begin{figure*}[t]
    \centering
    \includegraphics[width=1\linewidth]{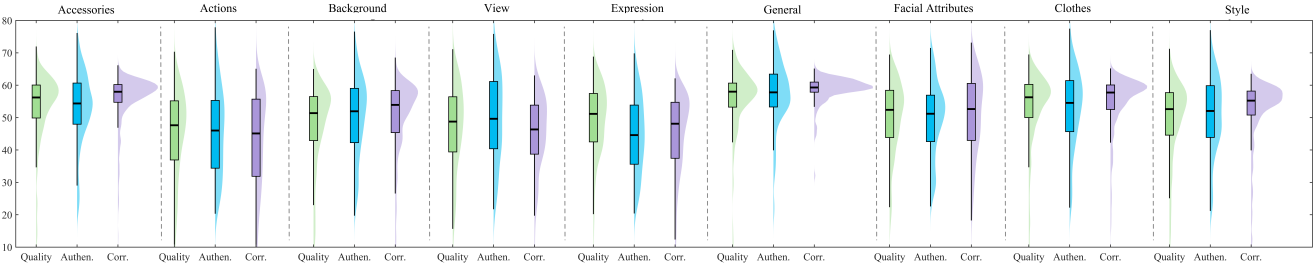}
    \vspace{-20pt}
    \caption{\textbf{Violin plots of quality, authenticity and correspondance scores in \textit{FaceQ-Gen} subset across nine prompt categories.}}
    \label{fig:prompt box}
    \vspace{-15pt}
\end{figure*}

\section{F-Bench: Benchmarking Current Generative Face Models}
In this section, we establish a human preference benchmark \textbf{F-Bench} for today's state-of-the-art face generation, customization, and restoration methods, from \textit{quality, authenticity, ID fidelity, and correspondence} dimensions. This benchmark analyzes the strengths and weaknesses, providing new insights and potential directions for future work.
\subsection{Perspective Analysis}
Fig.\ref{fig:mos} demonstrate the MOS distribution across eight dimensions in the three subsets, highlighting distinct variations between dimensions. (1) In the face generation task, the \textit{Correspondence} score is the highest among the three dimensions, indicating that existing generative models perform well in aligning with prompts. The distribution of \textit{Authenticity} is skewed towards lower scores, suggesting that current face generation models still fall short in generating photorealistic face images that satisfy human preferences. (2) In face customization, models perform the worst in the \textit{Quality} dimension overall, demonstrating the difficulty of injecting identity information without compromising image quality. \textit{Identity Fidelity} shows a bimodal distribution, reflecting the dichotomous nature of human judgment regarding whether the generated face accurately represents the same identity. (3) In face restoration, models perform better in \textit{Quality} than in \textit{Identity Fidelity}.

\subsection{Model-wise Comparison}
\begin{figure*}[t]
    \centering
    \includegraphics[width=1\linewidth]{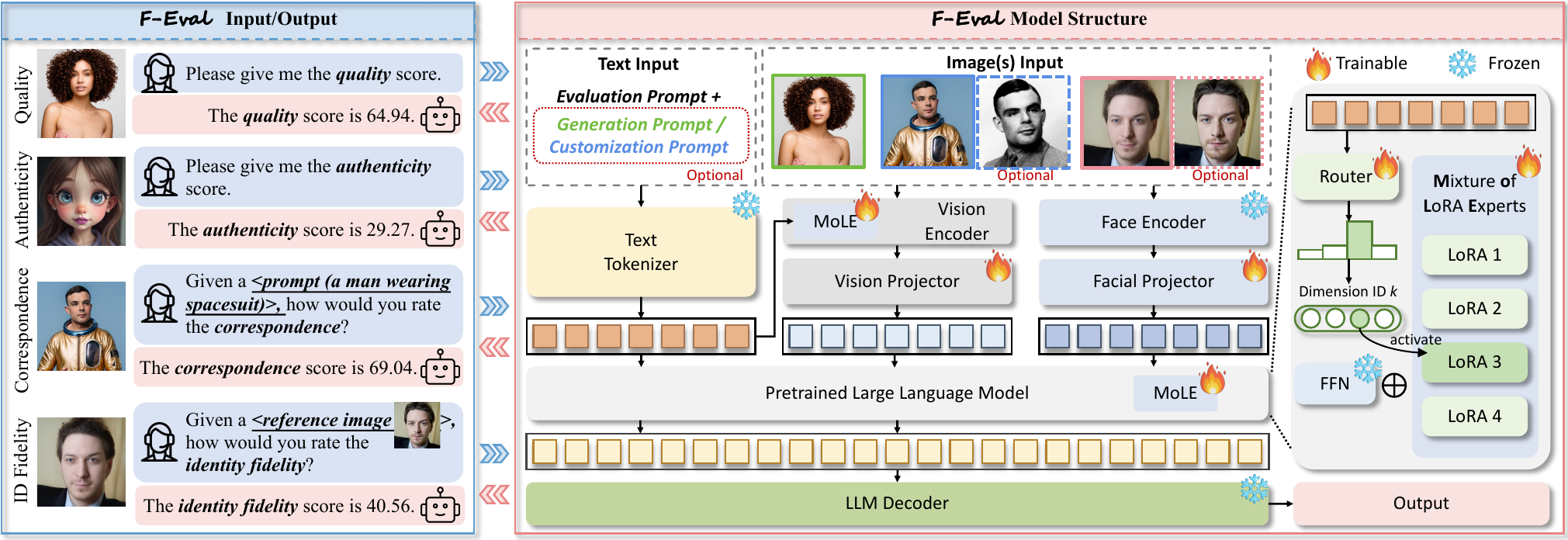}  
    \vspace{-20pt}
    \caption{\textbf{The overall framework of F-Eval}. F-Eval can evaluate quality, authenticity, correspondence, and identity fidelity in a \textbf{one-for-all} framework. It can process both single and paired images, along with prompts, to produce quality scores. It consists of three encoders, including a vision encoder, a face encoder, and a text tokenizer to process multi-modal inputs. These features are projected into the same space by trained projectors. A pre-trained large language model is utilized to fuse the features while fine-tuned with four LoRA experts. Specific LoRA will be activated by dimension ID, which is classified by a trainable router.}
    \label{fig:method}
    \vspace{-18pt}
\end{figure*}
\noindent\textbf{Face Generation.} Fig.\ref{fig:radar}\textcolor{iccvblue}{a} presents the average MOS scores across three dimensions: \textit{Quality}, \textit{Authenticity}, and \textit{Correspondence}. \textit{Correspondence} scores show a steady increase over time, indicating consistent improvement in text comprehension. \textit{Quality} scores generally rise with the release year, although Hunyuan~\cite{li2024hunyuanditpowerfulmultiresolutiondiffusion} and Dreamlike~\cite{dreamlike} exhibit deviations. Hunyuan~\cite{li2024hunyuanditpowerfulmultiresolutiondiffusion} oversmoothes the facial details and performs worse compared to its contemporaries, while Dreamlike~\cite{dreamlike} performs the opposite. \textit{Authenticity} scores exhibit the greatest variation among models, suggesting that current face generation models place insufficient emphasis on authenticity. Except for Realistic Vision~\cite{realistivision}, SD3~\cite{sd3}, Flux-dev~\cite{flux}, and Dreamlike~\cite{dreamlike}, most models lag in authenticity relative to quality, highlighting a common phenomenon of high-quality but less realistic outputs in generative models. Fig.\ref{fig:mos}\textcolor{iccvblue}{d,e}  give some examples of MOS distributions of different methods, revealing a reasonable fine-grained performance difference.\\
\noindent\textbf{Face Customization.} Fig.\ref{fig:radar}\textcolor{iccvblue}{b} is the performance radar chart for face customization models. IP-Adapter-FaceID-Plus\cite{ipa2} achieves the best performance across all dimensions, followed by PhotoMaker~\cite{li2023photomaker}. This highlights their great robustness in identity encoders. ELITE~\cite{wei2023eliteencodingvisualconcepts} scores the lowest in all dimensions. IP-Adapter-FaceID~\cite{ipa1} produces high-quality images but lacks accuracy in \textit{Identity Fidelity}. This may be because the base model~\cite{podell2023sdxlimprovinglatentdiffusion} introduces many AI artifacts, which affect both \textit{Identity Fidelity} and \textit{Quality}. InstantID~\cite{wang2024instantid} and FastComposer~\cite{xiao2023fastcomposertuningfreemultisubjectimage} score well in \textit{Identity Fidelity} but underperform significantly in \textit{Quality} and \textit{Correspondence}, indicating strong identity preservation but limited text-driven editing capabilities. Fig.\ref{fig:mos}\textcolor{iccvblue}{h} illustrate the model-wise MOS distributions, revealing similar trends to those observed in the radar chart.\\
\noindent\textbf{Face Restoration.} Fig.\ref{fig:radar}\textcolor{iccvblue}{c} displays the average performance comparison of nine face restoration models. DR2~\cite{wang2023dr2} exhibits the weakest performance in both \textit{Quality} and \textit{Identity fidelity}. This may be due to the degradation removal process of DR2~\cite{wang2023dr2}, which results in the loss of fine details and excessive smoothing. In contrast, StableSR~\cite{wang2024exploitingdiffusionpriorrealworld} and CodeFormer~\cite{zhou2022towards} achieve the best overall performance among the nine face restoration models, successfully preserving both intricate details and accurate identity representations. When comparing the performance on real-world and synthetic data, most methods struggle to achieve satisfactory results on complex real-world degradations, except for SAPRNet~\cite{chen2020learning} and DiffBIR~\cite{lin2023diffbir}. StableSR~\cite{wang2024exploitingdiffusionpriorrealworld} and GFPGAN~\cite{wang2021gfpgan} show strong robustness, performing consistently well on both real-world and synthetic data. 
\vspace{-0.5mm}
\subsection{Class-wise Comparison}
\vspace{-0.5mm}
For the face generation task, we analyze the Mean Opinion Score (MOS) distributions of \textit{Quality}, \textit{Authenticity}, and \textit{Correspondence} across nine prompt categories, as shown in Fig.\ref{fig:prompt box}. The violin plots highlight distinct performance patterns across these categories. The \textit{Action}, \textit{Expression}, and \textit{Facial Attributes} categories exhibit lower MOS values in all three dimensions, with notable variability. In the \textit{Action} category, limited visibility of the face due to body movements affects the generation of fine facial details, resulting in reduced scores, especially in \textit{Quality} and \textit{Correspondence}. The \textit{Expression} category poses challenges in maintaining spatial relationships and physical constraints, leading to distortion and artifacts in expressions with exaggerated facial movements, such as laughing or crying, thereby impacting both \textit{Quality} and \textit{Authenticity}. For \textit{Facial Attributes}, generating images with precise attributes (e.g., specific eye color, hair style) is challenging, particularly when multiple attributes are specified, which lowers the scores. In contrast, the \textit{General} and \textit{Background} categories achieve the highest MOS scores across all dimensions, as simpler or less specific prompts are easier for the model to interpret accurately. The \textit{Clothes} and \textit{Accessories} categories also perform well, indicating that the model handles clothes and accessory prompts effectively.
Overall, prompts with specific or complex attribute requirements tend to result in lower MOS scores, while general 
or simple prompts yield consistently high scores across dimensions.
This can be attributed to the absence of high-quality fine-grained text-face pair datasets during the training of generative models.

\begin{table*}[!t]
\centering
\huge
\caption{\textbf{Performance comparison on \textit{FaceQ-Gen} and \textit{FaceQ-Cus} subsets}. $\spadesuit$, $\clubsuit$, $\diamondsuit$, and $\heartsuit$ denote traditional IQA models, face IQA models, classical deep learning-based IQA models, and AIGC IQA models respectively. * indicates fine-tuned results.}
\vspace{-6pt}
\label{tab:face_generation}
\renewcommand{\arraystretch}{1.1} 
\resizebox{\textwidth}{!}{ 

\begin{tabular}{l|| c c c: c c c: c c c|| c c c: c c c: c c c}

\toprule[2pt]
\rowcolor{gray!20} 
Task &\multicolumn{3}{c:}{}&\multicolumn{3}{c:}{\textbf{Face Generation}} &\multicolumn{3}{c||}{}&\multicolumn{3}{c:}{}& \multicolumn{3}{c:}{\textbf{Face Customization}}&\multicolumn{3}{c}{}\\
\midrule
Dimension & \multicolumn{3}{c:}{Quality} & \multicolumn{3}{c:}{Authenticity} & \multicolumn{3}{c||}{Correspondence}& \multicolumn{3}{c:}{Quality} & \multicolumn{3}{c:}{ID Fidelity} & \multicolumn{3}{c}{Correspondence} \\
\cmidrule{2-4} \cmidrule{5-7} \cmidrule{8-10}\cmidrule{11-13}\cmidrule{14-16}\cmidrule{17-19}
Model & SRCC$\uparrow$ & KRCC$\uparrow$ & PLCC$\uparrow$ & SRCC$\uparrow$ & KRCC$\uparrow$ & PLCC$\uparrow$ & SRCC$\uparrow$ & KRCC$\uparrow$ & PLCC$\uparrow$ & SRCC$\uparrow$ & KRCC$\uparrow$ & PLCC$\uparrow$ & SRCC$\uparrow$ & KRCC$\uparrow$ & PLCC$\uparrow$ & SRCC$\uparrow$ & KRCC$\uparrow$ & PLCC$\uparrow$ \\
\midrule
$\spadesuit$NIQE \cite{niqe} & 0.1726 & 0.1142 & 0.2553 & 0.0869 & 0.0571 & 0.1346 & 0.0752 & 0.0507 & 0.1098&0.1365&0.0899&0.1436&0.1490&0.0991&0.1310&0.1186&0.0784&0.1245 \\
$\spadesuit$ILNIQE \cite{ilniqe} & 0.1981 & 0.1319 & 0.3159 & 0.0675 & 0.0445 & 0.1505 & 0.0889 & 0.0591 & 0.1665&0.2032&0.1327&0.2479&0.2192&0.1452&0.2778&0.1878&0.1251&0.2275 \\
$\spadesuit$HOSA \cite{hosa} & 0.2537 & 0.1708 & 0.2972 & 0.1103 & 0.0728 & 0.1446 & 0.122 & 0.0814 & 0.1495&0.1019&0.0688&0.0597&0.0753&0.0506&0.0552&0.1036&0.0696&0.0748 \\
$\spadesuit$BPRI-PSS \cite{pri} & 0.2362 & 0.1595 & 0.2413 & 0.1700 & 0.1142 & 0.1772 & 0.1673 & 0.1121 & 0.1647 &0.0238&0.0166&0.0314&0.1425&0.0972&0.1758&0.1100&0.0729&0.1067\\
$\spadesuit$FISBLIM \cite{fisblim} & 0.0901 & 0.0595 & 0.0433 & 0.0028 & 0.0003 & 0.0175 & 0.1000 & 0.0665 & 0.0196 &0.1614&0.1024&0.1690&0.0256&0.0173&0.0126&0.0852&0.0559&0.0854\\
\midrule 
$\clubsuit$ArcFace\cite{deng2019arcface}&N/A&N/A&N/A&N/A&N/A&N/A&N/A&N/A&N/A&0.3099&0.2094&0.3605&0.5062&0.3439&0.5572&0.1837&0.1243&0.2244\\
$\clubsuit$SER-FIQ\cite{ser-fiq} & 0.1552 & 0.1034 & 0.1561 &0.0470 & 0.0314 & 0.0575 & 0.1130 & 0.075 & 0.1110 & -0.1241 & -0.080 & -0.1241 & -0.0109 & -0.0078 & 0.0042 & -0.1680 & -0.1121 & -0.1911  \\
 $\clubsuit$DSL-FIQA\cite{dsl-fiqa} & 0.5606 & 0.3927& 0.5945 & 0.3156 & 0.2096 & 0.3960 & 0.3706 & 0.2514 & 0.4355  & 0.3615 &0.2450 & 0.3651 & 0.2429 & 0.1652 & 0.2519 & 0.1040 & 0.0703 & 0.0974\\
  $\clubsuit$IFQA\cite{ifqa} & 0.2313 &0.1549 & 0.3261& 0.1597 & 0.1062 & 0.2097&0.1597 & 0.1077 &  0.2260 & 0.2401 & 0.1614 &0.2220 &0.1503 &0.0973 & 0.1739& 0.3731 &0.2546 &0.3941 \\
\midrule
$\diamondsuit$CNNIQA\textsuperscript{*} \cite{cnniqa} & 0.4219 & 0.2895 & 0.3510 & 0.3037 & 0.2054 & 0.2973 & 0.2824 & 0.1931 & 0.2595&0.6244&0.4376&0.6620&0.6511&0.4441&0.6381&0.3765&0.2545&0.3264 \\
$\diamondsuit$VGG16\textsuperscript{*} \cite{vgg} &0.5846 & 0.4141 & 0.6006& 0.5010 & 0.3508 & 0.5181 & 0.4349 & 0.2994 & 0.4723&0.7217&0.5324&0.7857&0.6735&0.4717&0.6636&0.5635&0.3932&0.5589 \\
$\diamondsuit$VGG19\textsuperscript{*} \cite{vgg} & 0.5728 & 0.4043 & 0.5644 & 0.4681 & 0.3248 & 0.4479 & 0.4171 & 0.2873 & 0.4754&0.7952&0.6031&0.8096&0.7544&0.5417&0.7555&0.6585&0.4742&0.6465 \\
$\diamondsuit$ResNet18\textsuperscript{*} \cite{resnet} & 0.6150& 0.4417 &0.6660 & 0.4600 &0.3160 &0.4604 & 0.4829 & 0.3360 & 0.4628 &0.7966 & 0.6013&0.8006 & 0.7473 & 0.5335 &0.7656&0.6308&0.4488 & 0.6191\\
$\diamondsuit$ResNet34\textsuperscript{*} \cite{resnet}& 0.6092 & 0.4354 & 0.6537 & 0.5564 & 0.3949 & 0.5401 & 0.4410 & 0.3065 & 0.5207&0.8141 & 0.6211 &0.8151&0.7969&0.5878 & 0.8157 & 0.5708 &0.4000& 0.5810\\
 $\diamondsuit$HyperIQA\textsuperscript{*} \cite{hyperiqa} & 0.6333 & 0.4545 & 0.6591 & 0.6106 & 0.4406 & 0.6042 & 0.4235 & 0.2912 & 0.4240&{0.8524}&0.6631&0.8205&0.8419&0.6463&0.8588&0.6628&0.4757&0.6454 \\
$\diamondsuit$TReS\textsuperscript{*} \cite{tres} & 0.7766 & 0.5867 & 0.8047 & 0.6458 & 0.4703 & 0.6579 & 0.5708 & 0.4038 & 0.6270 &
{0.8883}&0.7083&{0.8951}&0.8330&0.6409&0.8658&0.7393&0.5497&0.7400\\
$\diamondsuit$MANIQA\textsuperscript{*} \cite{maniqa} &{0.7871} & {0.5951} & {0.8150} & {0.7308} & {0.5436} &{0.7794} & {0.6278} & {0.4443} & {0.7066}&0.8970&0.7185&0.8952&0.8590&{0.6654}&{0.8796}&{0.7849}&{0.5941}&{0.7822} \\
$\diamondsuit$PromptIQA\textsuperscript{*} \cite{chen2024promptiqa} &0.7823 & 0.5907 & 0.8245  & 
0.7232 & 0.5327 & 0.7558 &
0.6514& 0.4698&0.7065&\textcolor{blue}{0.9071}& \textcolor{blue}{0.7315}&\textcolor{blue}{0.8821}&
\textcolor{blue}{0.8693}& \textcolor{blue}{0.6792}&\textcolor{blue}{0.8876}&
0.8348&0.6342& 0.8281
 \\

\midrule
 $\heartsuit$CLIPScore \cite{clipscore} & 0.0967 & 0.0642 & 0.1315 & 0.0640 & 0.0408 &0.0817 & 0.2396 & 0.1591 & 0.2733&0.3126&0.2106&0.3185&0.1816&0.1218&0.1811&0.7008&0.5098&0.6986 \\
 $\heartsuit$BLIPScore \cite{blip}& 0.1306 & 0.0872 & 0.1929 & 0.0935 & 0.0613 & 0.1569 & 0.2051 & 0.1364 & 0.2963 &0.3073&0.2063&0.3136&0.1213&0.0796&0.1266&0.6707&0.4797&0.6123\\
$\heartsuit$ImageReward \cite{imagereward}& 0.3849 & 0.2631 & 0.4358 & 0.2606 & 0.1736 & 0.3181 & 0.5155 & 0.3542 & 0.5871  & 0.3232 & 0.2166 & 0.3196 & 0.1905 & 0.1270 & 0.1839 & 0.7714 & 0.5673 & 0.7740\\
$\heartsuit$MINTIQA\cite{mintiqa}&0.7002
&0.5160
&0.8047
&0.6217
&0.4450
&0.6826
&0.5491
& 0.3891
& 0.6584&0.5378
&0.3775
&0.5574
&0.4088
&0.2816
&0.4041
&0.2886
&0.1932
&0.2959\\
$\heartsuit$MINTIQA\textsuperscript{*} \cite{mintiqa}& \textcolor{blue}{0.8312} & \textcolor{blue}{0.6474} & \textcolor{blue}{0.8974} & \textcolor{blue}{0.8177} & \textcolor{blue}{0.6306} & \textcolor{blue}{0.8511} & \textcolor{blue}{0.7908} & \textcolor{blue}{0.5991} & \textcolor{blue}{0.8667}&{0.8904}&{0.7092}&{0.8816}&{0.8524}&{0.6697}&{0.8755}&\textcolor{blue}{0.8391}&\textcolor{blue}{0.6413}&\textcolor{blue}{0.8318} \\

\midrule
\textbf{F-Eval (Ours)} & \textcolor{red}{0.8486} & \textcolor{red}{0.6670} & \textcolor{red}{0.9085} & \textcolor{red}{0.8312} & \textcolor{red}{0.6585} & \textcolor{red}{0.8578} & \textcolor{red}{0.8471} & \textcolor{red}{0.6637} & \textcolor{red}{0.9106}&\textcolor{red}{0.9462}&\textcolor{red}{0.7961}&\textcolor{red}{0.9461}&\textcolor{red}{0.9188}&\textcolor{red}{0.7640}&\textcolor{red}{0.9322}&\textcolor{red}{0.9460}&\textcolor{red}{0.7959}&\textcolor{red}{0.9457} \\
\bottomrule[2pt]
\end{tabular}}
\vspace{-17pt}
\end{table*}

\section{F-Eval Method}
In this section, we introduce our \textbf{one-for-all} generated face quality assessment method, \textbf{F-Eval}, towards predicting quality scores from multiple dimensions for three tasks using one model.
\subsection{Overall Architecture}
The overall architecture of F-Eval is depicted in Fig.~\ref{fig:method}. As shown on the left,
F-Eval can assess four dimensions: quality, authenticity, text-image correspondence, and identity fidelity, with multimodal inputs, \textit{i.e.}text prompts, image and optional reference image, to output quality-related answers. On the right, the input image(s) can originate from all three tasks. F-Eval employs a vision encoder and a face encoder to extract visual features, with projectors projecting both into a unified language space. User prompts are encoded into textual embeddings through a tokenizer. These multimodal features are concatenated and fed into a pre-trained large language model for the final output.

\subsection{Model Design}

\paragraph{Vision Encoding.} For input image(s), F-Eval adopt a vision encoder to extract global features and a face encoder captures face features related to facial structure and identity. Specifically, the vision encoder, Qwen-VL-2.5-ViT~\cite{bai2025qwen2}, is tuned on FaceQ to adjust different dimensions. Face encoder is Arcface~\cite{deng2019arcface}, which is frozen. Two two-layer multi-layer perception projectors then map the features .\\
\paragraph{Feature Fusion via the LLM.}
For input prompts, it is tokenized into text tokens and then concatenated with visual tokens. Here, Qwen-VL2.5-7B~\cite{bai2025qwen2} is employed as the pre-trained LLM backbone to combine visual tokens and text tokens to perform multi-modal learning.. Finally, the output features of the LLM are decoded with a text decoder and then projected to the prompt-corresponded quality space.
\vspace{-0.3mm}
\subsection{Multi-Expert LoRA Tuning}
\vspace{-0.3mm}
\paragraph{Instruction Tuning.} 
Previous IQA methods are often designed for specific dimensions~\cite{maniqa,zhang2018lpips}, addressing only one aspect of the problem or requiring retraining of model weights for different dimensions. Leveraging the generalization ability of LLMs, instruction tuning allows for the integration of different dimensions within a unified framework. Therefore, we construct diverse instruction data for the four facial quality evaluation dimensions in the Face-Q dataset and fine-tune F-Eval by adjusting the visual encoder, projector, and LLM parameters to adapt it to the specific task. As a result, F-Eval is capable of evaluating facial quality using unified weights.

\paragraph{Mixture of LoRA Experts (MoLE).} Tuning a holistic quality model for different dimensions using a single LoRA expert introduces inherent interference between dimensions. Hence, F-Eval integrates four LoRA experts for four dimensions when tuning vision encoder and LLM, constituting MoLE modules. A router is trained to classify input prompts into the four dimensions and output the corresponding dimension ID $k$.    Specifically, for a frozen linear layer \( h = Wx \) with input \( x \in \mathbb{R}^{d_i} \) and weight matrix \( W \in \mathbb{R}^{d_o \times d_i} \), a LoRA expert is trained to learn a low-rank decomposed update:
\begin{equation}
    h = Wx + \Delta Wx = Wx + \frac{\alpha}{r} BAx,
\end{equation}
where \( A \in \mathbb{R}^{r \times d_i} \) and \( B \in \mathbb{R}^{d_o \times r} \) are the low-rank matrices, \( r \ll \min(d_o, d_i) \) is the rank, and \( \alpha \) is scaling factor.
 For prompt of certain dimension, the corresponding expert is activated based on the dimension ID $k$, which can be formulated as:
 \begin{equation}
    E_k(x)=\frac{\alpha_k}{r_k} B_kA_kx,
\end{equation}
 where \(E_k(\cdot)\) is the k-th LoRA expert.

\section{Experiments}

\subsection{Implementation Details}
The LLM used in F-Eval is Qwen-2.5VL~\cite{bai2025qwen2}, with an input token channel dimension of 3584. The vision encoder, face encoder, and LLM are all frozen during training. The ranks of LoRA experts are 16,32,16,16, respectively. The LoRA alphas are set equal to LoRA ranks. LoRA dropout is set as 0.05. The model is trained on 2 NVIDIA GeForce RTX 3090 GPUs with flash-attn~\cite{dao2023flashattention2fasterattentionbetter} enabled. We use AdamW Optimizer to train the network for 10 epochs with a batch size of 4 and learning rate $1e^{-4}$. 
\vspace{-1mm}
\subsection {Comparison with existing QA Models}
We conduct comprehensive experiments on the proposed FaceQ database to evaluate the effectiveness of both current quality assessment methods and our proposed methods, as shown in Tab.\ref{tab:face_generation}, Tab.\ref{face res syn}, and Tab.\ref{face res rw}. The evaluation metrics are  Spearman rank-order correlation coefficient (SRCC), Kendall rank order correlation coefficient (KRCC), and Pearson linear
correlation coefficient (PLCC), which assess the correlation between predicted scores by the QA methods and the MOS scores provided by humans. Higher metrics indicated better alignment with human preferences.
The comparison QA models falls into four categories:
1) Traditional IQA Models. NR-IQA models include NIQE \cite{niqe}, ILNIQE \cite{ilniqe}, HOSA \cite{hosa}, BPRI-PSS \cite{pri}, FISBLIM \cite{fisblim}. FR-IQA models include LPIPS \cite{zhang2018lpips},
SSIM \cite{wang2004image} and PSNR \cite{psnr}. 2) Classical Deep learning-based IQA Models. This category includes CNNIQA \cite{cnniqa}, VGG16 \cite{vgg}, VGG19 \cite{vgg}, ResNet18 \cite{resnet}, ResNet34 \cite{resnet}, HyperIQA ~\cite{hyperiqa}, TReS \cite{tres}, MANIQA \cite{maniqa},PromptIQA\cite{chen2024promptiqa}. All models are retrained on FaceQ with an 80:20 training/testing split. 3) AIGC IQA Models. This category includes pre-trained vision-language models, including CLIPScore \cite{clipscore}, BLIPScore \cite{blip}, zero-shot metric ImageReward \cite{imagereward}, and the AIGC-specific QA model MINTIQA \cite{mintiqa}. 4) Face IQA Models. This category includes SER-FIQ \cite{ser-fiq}, DSL-FIQA \cite{dsl-fiqa}, ArcFace \cite{deng2019arcface}, IFQA\cite{ifqa}.

\vspace{-1mm}
\subsection{Performance Analysis}
\begin{table}[!t]
\centering
\vspace{0mm}
\caption{\textbf{Performance comparison on \textit{FaceQ-Res} synthetic subset.} $\spadesuit$, $\heartsuit$, $\clubsuit$, and $\diamondsuit$ denote traditional NR-IQA models, FR-IQA models, face IQA models, and classical deep learning-based IQA models respectively. * indicates fine-tuned results.}
\vspace{-3mm}
\label{face res syn}
\renewcommand{\arraystretch}{0.7} 
\resizebox{\columnwidth}{!}{  

\begin{tabular}{l|| c c c : c c c }
\toprule
Dimension & \multicolumn{3}{c:}{Quality} & \multicolumn{3}{c}{ID Fidelity} \\
\cmidrule{2-4} \cmidrule{5-7} 
Model&SRCC$\uparrow$&KRCC$\uparrow$&PLCC$\uparrow$&SRCC$\uparrow$&KRCC$\uparrow$&PLCC$\uparrow$\\
\midrule

$\spadesuit$BMPRI\cite{bmpri} &0.2161&0.1428&0.1862&0.1113&0.0739&0.1069\\
$\spadesuit$NIQE\cite{niqe} & 0.1798&0.1213&0.1642&0.2193&0.1480&0.2118\\
$\spadesuit$ILNIQE\cite{ilniqe} &0.2505&0.1727&0.3180&0.2696&0.1857&0.3321\\
$\spadesuit$HOSA\cite{hosa} &0.3758&0.2522&0.3748&0.3418&0.3455&0.2301\\
$\spadesuit$BPRI-LSSn \cite{pri}&0.1398
&0.0957
&0.0999
&0.1244
&0.0837
&0.0973
\\
$\spadesuit$BPRI \cite{pri}&0.1637
&0.1065
&0.1480
&0.0195
&0.0121
&0.0281
\\
$\spadesuit$FISBLIM \cite{fisblim}&0.1589&0.1089&0.2038&0.1054&	0.0709&0.1715\\
\midrule
$\heartsuit$LPIPS\cite{zhang2018lpips}&0.3909&0.2658&0.4643&0.4372&0.3018&0.4870\\
$\heartsuit$SSIM\cite{wang2004image}&0.1017&0.0675&0.0488&0.0933&0.0601&0.0504\\
$\heartsuit$PSNR \cite{psnr} &0.0198&0.0113&0.0536&0.0986&0.0662&0.1212\\
\midrule
$\clubsuit$ArcFace\cite{deng2019arcface}&0.3466&0.2327&0.3774&0.5728&0.4030&0.5762\\
$\clubsuit$SER-FIQ \cite{tres}&0.3813&0.2584&0.3746&0.3951&0.2641&0.3707\\
$\clubsuit$DSL-FIQA \cite{dsl-fiqa}&0.6387&0.4508&0.6528&0.4882&0.3360&0.5379\\
$\clubsuit$IFQA \cite{ifqa}&0.2273 & 0.1527 & 0.3348&0.3214 & 0.2185 & 0.3874\\
\midrule
$\diamondsuit$CNNIQA\textsuperscript{*} \cite{cnniqa}&0.4661&0.3197&0.3836&0.4678&0.3216&0.4288\\
$\diamondsuit$ResNet18\textsuperscript{*} \cite{resnet}&0.7254&0.5295&0.6982&0.6662&0.4802&0.6527\\
$\diamondsuit$ResNet34\textsuperscript{*}\cite{resnet}&0.6693&0.4916&0.6550&0.5905&0.4170&0.6084\\
$\diamondsuit${VGG16}\textsuperscript{*} \cite{vgg}&0.7487&0.5513&0.7101&0.6177&0.4413&0.6381\\
$\diamondsuit${VGG19}\textsuperscript{*} \cite{vgg}&0.6994&0.5126&0.6722&0.6038&0.4278&0.5999\\
$\diamondsuit$HyperIQA\textsuperscript{*} \cite{hyperiqa}&0.8269&0.6323&0.8228&0.7615&0.5585&0.7471\\
$\diamondsuit$MANIQA\textsuperscript{*} \cite{maniqa}&{0.8287}&{0.6403}&\textcolor{blue}{0.8778}&{0.7704}&{0.5660}&{0.7619}\\
$\diamondsuit$TReS\textsuperscript{*} \cite{tres}&\textcolor{blue}{0.8656}&\textcolor{blue}{0.6830}&{0.8622}&{0.7951}&{0.5879}&{0.7829}\\
$\diamondsuit$PromptIQA\textsuperscript{*} \cite{chen2024promptiqa}&0.8519& 0.6661&0.8733&
\textcolor{blue}{0.8220}& \textcolor{blue}{0.6275}&\textcolor{blue}{0.8254}\\
\midrule
\textbf{F-Eval (Ours)} &\textcolor{red}{0.8692}&\textcolor{red}{0.6855}&\textcolor{red}{0.9009}&\textcolor{red}{0.8507}&\textcolor{red}{0.6731}&\textcolor{red}{0.8726}\\
\bottomrule
\end{tabular}
}
  \vspace{-2mm}
\end{table}

\begin{table}[!t]
\centering
\vspace{-2mm}
\caption{\textbf{Performance comparison on \textit{FaceQ-Res} real-world subset.} $\spadesuit$, $\clubsuit$, and $\diamondsuit$ denote traditional IQA models, face IQA models, and classical deep learning-based IQA models respectively. * indicates fine-tuned results.}
\vspace{-3mm}
\label{face res rw}
\renewcommand{\arraystretch}{0.7} 
\resizebox{\columnwidth}{!}{  
\begin{tabular}{l|| c c c: c c c }
\toprule

Dimension & \multicolumn{3}{c:}{Quality} & \multicolumn{3}{c}{ID Fidelity} \\
  \cmidrule{2-4} \cmidrule{5-7} 
Model&SRCC$\uparrow$&KRCC$\uparrow$&PLCC$\uparrow$&SRCC$\uparrow$&KRCC$\uparrow$&PLCC$\uparrow$\\
\midrule
$\spadesuit$BMPRI\cite{bmpri} &0.1582&0.1093&0.2799&0.0149&0.0108&0.1386\\
$\spadesuit$NIQE\cite{niqe} &0.1783&0.1186&0.1962&0.1245&0.0810&0.1701\\
$\spadesuit$ILNIQE\cite{ilniqe} &0.2230&0.1610&0.2291&0.1964&0.1362&0.2340\\
$\spadesuit$HOSA\cite{hosa} &0.3927&0.4277&0.2654&0.3328&0.2234&0.3813\\
$\spadesuit$BPRI-LSSn \cite{pri}&0.2715&0.1840&0.3022&0.2059&0.1380&0.2475\\
$\spadesuit$BPRI \cite{pri}&0.1920&0.1313&0.2200&0.0846&0.0570&0.1136\\
$\spadesuit$FISBLIM \cite{fisblim}&0.2512&0.1705&0.2725&0.2341&0.1581&0.2699\\
\midrule
$\clubsuit$ArcFace\cite{deng2019arcface}&0.2371&0.1565&0.4183&0.4956&0.3470&0.6330\\
$\clubsuit$SER-FIQ \cite{tres}&-0.0482&-0.0299&0.1023&0.0421&0.0299&0.1584\\
$\clubsuit$DSL-FIQA \cite{dsl-fiqa}&0.6509&0.4641&0.6494&0.4840&0.3362&0.5118\\
$\clubsuit$IFQA \cite{ifqa}&0.0251 & 0.0210 & 0.1221&0.0893 & 0.0603 & 0.2086\\
\midrule
$\diamondsuit$CNNIQA\textsuperscript{*} \cite{cnniqa}&0.3650&0.2506&0.4705&0.3252&0.2197&0.3112\\
$\diamondsuit$VGG16\textsuperscript{*}\cite{vgg}&0.4996&0.3439&0.4269&0.4117&0.2859&0.4622\\
$\diamondsuit$VGG19\textsuperscript{*} \cite{vgg}&0.5871&0.4181&0.5133&0.3054&0.2044&0.3594\\
$\diamondsuit$ResNet18\textsuperscript{*} \cite{resnet}&0.5977&0.4221&0.5722&0.5554&0.3909&0.6161\\
$\diamondsuit$ResNet34\textsuperscript{*} \cite{resnet}&0.6259&0.4475&0.6168&0.5036&0.3539&0.6136\\
$\diamondsuit$HyperIQA\textsuperscript{*} \cite{hyperiqa}&{0.7926}&0.5965&0.7926&\textcolor{blue}{0.7783}&\textcolor{blue}{0.5845}&\textcolor{blue}{0.8163}\\
$\diamondsuit$MANIQA\textsuperscript{*} \cite{maniqa}&\textcolor{blue}{0.8356}&\textcolor{blue}{0.6482}&{0.8467}&{0.7597}&{0.5655}&{0.8081}\\
$\diamondsuit$TReS\textsuperscript{*} \cite{tres}&{0.8287}&{0.6391}&{0.8501}&0.7287&0.5441&0.8009\\
$\diamondsuit$PromptIQA\textsuperscript{*} \cite{chen2024promptiqa}&0.8115& 0.6186&\textcolor{blue}{0.8645}& 
0.7398& 0.5526&0.8297\\
\midrule
\textbf{F-Eval(Ours)} &\textcolor{red}{0.8448}&\textcolor{red}{0.6577}&\textcolor{red}{0.8705}&\textcolor{red}{0.7957}&\textcolor{red}{0.6057}&\textcolor{red}{0.8366}\\
\bottomrule
\end{tabular}
}
  \vspace{-5mm}
\end{table}
\paragraph{Traditional IQA Models.} Traditional IQA methods, including both NR-IQA and FR-IQA models, generally perform poorly across all tasks and dimensions, with low SRCC, KRCC, and PLCC scores. \\
\paragraph{Classical Deep Learning-based IQA Models.} 
After being trained on our dataset, deep learning-based IQA models, such as MANIQA~\cite{maniqa} and HyperIQA~\cite{hyperiqa}, significantly outperform traditional models in terms of quality and authenticity. However, these methods face challenges in text understanding, leading to unsatisfactory correlations.\\
\paragraph{AIGC IQA Models.} CLIPScore~\cite{clipscore} BLIPScore~\cite{blip}, and ImageReward~\cite{imagereward} exhibit weak correlations with human preference, likely due to their focus on high-level semantics rather than low-level details crucial for image quality assessment. MINTIQA~\cite{mintiqa} achieves best performance among current AIGC IQA methods, but its performance is still unsatisfactory, particularly on the customization task. Fine-tuning on our dataset leads to significant improvements, highlighting the great potential of FaceQ.\\
\paragraph{Face IQA Models.} ArcFace similarity~\cite{deng2019arcface} demonstrates its ability to extract identity-related features. However, it lacks perceptual sensitivity for image quality, leading to poor performance in quality assessment. DSL-FIQA~\cite{dsl-fiqa}, as the latest method, achieves the best performance in predicting quality scores among face IQA methods, but it is still far from satisfactory. SER-FIQ fails to align with human perception across most dimensions and tasks. Experiments confirm that these natural face evaluation algorithms struggle to assess the complex characteristics of AIGFs, highlighting the need for an effective AIGF evaluator.\\
\paragraph{F-Eval (Ours).} 
Our method achieves optimal performance across all tasks and dimensions, with the most significant gain in the customization task, as existing methods rarely focus human preferences in face customization. Among the four dimensions, the improvements in correspondence and ID fidelity stand out, thanks to dual-stream vision encoding and the powerful LLM backbone for text understanding and modality fusion. The MoLE technique allows the model to learn distinct features for each dimension.
\vspace{-1mm}
\subsection{Ablation Study}
We conducted the ablation study to validate the utility of the core components of F-Eval in Tab.~\ref{ablation}.\\
\paragraph{Effectiveness of Vision Encoding.} 1) The effectiveness of the face encoder is demonstrated by comparing the second and last rows. Without the face encoder, a significant performance drop is observed, indicating that the extracted features help the LLM interpret face images. 2) Removing the projector leads to a dramatic performance drop.\\
\paragraph{Effectiveness of MoLE.} We conducted three sets of LoRA adaptation experiments: fine-tuning a single LoRA for all dimensions and tasks, fine-tuning three LoRA experts for three tasks, and fine-tuning four LoRA experts for four dimensions. The results show that fine-tuning multiple dimensions with a single LoRA leads to performance deterioration, which further confirms the independence of the characteristics across different dimensions.

\begin{table}[h]
\vspace{-2mm}
\renewcommand\arraystretch{1.2}
  \caption{Ablation study of the proposed F-Eval method. Here only present SRCC due to space limitations.}

  \vspace{-3mm}
  \resizebox{1\columnwidth}{!}{
  \begin{tabular}{cccccc| ccc ccccccc}
    \toprule
                                    \multicolumn{6}{c}{Feature \& Strategy}                                  & \multicolumn{3}{c}{Face Generation}                   & \multicolumn{3}{c}{Face Customization}                  & \multicolumn{2}{c}{Face Restoration syn} &
                                   \multicolumn{2}{c}{Face Restoration rw} \\
    \cmidrule(r){1-6} \cmidrule(r){7-9} \cmidrule(r){10-12} \cmidrule(r){13-14}\cmidrule(r){15-16}
      Vision      & Projector  & Face   & Single LoRA & Task LoRA     & Dim. LoRA& Quality     & Authen.     & Corres.          & Quality       & ID.    & Corres.          & Quality     & ID.   & Quality          & ID.        \\
      \ding{52}     &         &   \ding{52}  &  &           & \ding{52}       & 0.808     & 0.750     & 0.846      & 0.924    &  0.843    & 0.927     &0.777     & 0.776     & 0.707    & 0.478       \\
    
         \ding{52}     &    \ding{52}     &     &   &          & \ding{52}       &   0.815   &  0.752    &  0.835     &  0.924  &   0.855   &  0.921   &   0.812 & 0.783    & 0.716   &  0.603        \\

    \ding{52}          &    \ding{52} &    \ding{52}    &  \ding{52}         &  &              & 0.793     & 0.776    & 0.819     & 0.941    &   0.885  & 0.942    & 0.838     & 0.823     & 0.802     & 0.671     \\
    
    \ding{52}          & \ding{52} & \ding{52} &  & \ding{52}          &           & 0.823     & 0.780     & 0.838     & 0.942     &   0.873   & 0.942     & 0.824     & 0.826     & 0.754     & 0.638      \\    
   \rowcolor{gray!20} \ding{52}   &\ding{52}       & \ding{52}   &  &  & \ding{52}  &\textbf{ 0.849 }    & \textbf{0.831 }     & \textbf{0.847 }    &\textbf{0.946 }    & \textbf{0.919 }    & \textbf{0.946 }   & \textbf{0.869     }& \textbf{0.851 }   & \textbf{0.845 }    & \textbf{0.796 }      \\
    \bottomrule
  \end{tabular}\label{ablation}
   }
  \vspace{-5mm}
  \centering
\end{table}
\label{sec:exp}

\section{Conclusion}
\label{sec:conclusion}
In this work, we conduct a pioneering generative face image quality assessment study. Specifically, we introduce FaceQ, the first generated face quality assessment database comprising 12,255 face images and 491,130 human preference ratings. Then, we establish F-Bench, a benchmark for evaluating the capabilities of existing face generation, customization, and restoration models, while also reveals the limitations of current Image Quality Assessment (IQA) and Face Quality Assessment (FQA) methods. Furthermore, we propose an evaluation model F-Eval to accurately predict human preferences on generated face images, which demonstrates state-of-the-art performance. We believe this work will contribute to advancing generative models and help overcome current limitations in AIGC community.

\section*{Acknowledgment}
This work is supported by the National Natural Science Foundation of China 
under Grants 62271308, 62401365, 62225112, 62132006, U24A20220, 
STCSM under Grants 24ZR1432000, 24511106902, 24511106900, 22511105700, 22DZ2229005, 
111 plan under Grant Number BP0719010, China Postdoctoral Science Foundation 
under Grants BX20250411, 2025M773473, 
and State Key Laboratory of UHD Video and Audio Production and Presentation.

{
    \small
    \bibliographystyle{ieeenat_fullname}
    \bibliography{main}
}
\clearpage
\setcounter{page}{1}
\maketitlesupplementary

\paragraph{Limitations and Social Impact.}
First, we are mindful of privacy concerns related to face datasets, and all data collection and sharing adhere to relevant privacy policies. Second, while the maximum resolution in our dataset is limited to 1024 pixels, we recognize that future work may incorporate higher-resolution images as generative models continue to advance. 
\section{\textit{FaceQ}: Dataset Construction}
\subsection{Face Generation, Customization and Restoration Models}
\paragraph{Model Implementation Details.} Tab.~\ref{table1} provides a comprehensive summary of models evaluated for face generation, editing, and restoration, including the model links, released dates, the resolution, and the backbone architectures. 
(1) Face generation. All the generation models are inference by pre-trained checkpoints in their default resolutions and hyper-parameters. Specifically, Stable Diffusion V1.5~\cite{Rombach_2022_CVPR}, DreamLike~\cite{dreamlike}, and RealisticVision~\cite{realistivision} support high-resolution generation, such as 1024$\times$ 1024, but we utilize their default training resolution due to severe subject repetition phenomenon. Deep Floyd \cite{IF} is a 3-stage pixel space diffusion model, here we only consider the third-stage results. For the dynamic step sampling, the number of steps per stage for StableCascade~\cite{stablecascade} and Deep Floyd~\cite{IF} is reduced to the quarter. Each model’s negative prompt is configured to exclude “anime” and “semi-realistic” outputs by using terms like “worst quality”, “low quality”, “illustration”, “3D”, “2D”, “painting”, “cartoons”, “sketch”, “anime”, “animation”, “cartoon”, and “semi-realistic”. Safe sensors are enabled to filter out NSFW content.
(2) Face customization. All the customization models are inference by pre-trained checkpoints in their default resolutions and hyper-parameters. FastComposer~\cite{xiao2023fastcomposertuningfreemultisubjectimage}, originally designed for multi-subject customization, is assessed here with a single reference image as input. We use the IP-Adapter release version, including IP-Adapter-FaceID-SDXL~\cite{ipa1} and IP-Adapter-FaceID-PlusV2~\cite{ipa2} (referred to as IP-Adapter-FaceID and IP-Adapter-FaceID-Plus). Their backbones are SDXL and SD-v1.5, respectively.
(3) Face restoration. For DR2~\cite{wang2023dr2}, we follow the hyperparameter settings recommended in the original paper:
$N=4, T=35$ for real-world inputs and $N=8, T=35$ for synthetic inputs.
\begin{table}[ht]
    \centering
    \caption{\textbf{Summary of 29 face generation, customization and restoration models.}}
    \label{table1}
    \resizebox{\columnwidth}{!}{  
    \begin{tabular}{@{}lllll@{}}
        \toprule[1pt]
        \textbf{\textit{Category}} &\textbf{Model} & \textbf{Year} & \textbf{Resol.} & \textbf{Backbone}\\ 
        \midrule
        \multirow{14}{1.7cm}{\centering \textit{Face \\Generation}}& 
        
        Stable Diffusion V1.5 ~\cite{rombach2022highresolutionimagesynthesislatent}& 2022.04 &$512^2$& Latent Diffusion\\
        &Stable Diffusion V2.1 ~\cite{rombach2022highresolutionimagesynthesislatent}& 2022.12 &  $1024^2$ & Latent Diffusion\\ 
        &DreamLike V2.0~\cite{dreamlike}& 2023.01 & $768^2$ & Latent Diffusion\\
        &Deep Floyd \cite{IF}& 2023.04 & $1024^2$ & Pixel Diffusion\\
        &SD-XL ~\cite{podell2023sdxlimprovinglatentdiffusion}& 2023.06 & $1024^2$ & Latent Diffusion\\
        &PixArt-alpha~\cite{chen2023pixartalphafasttrainingdiffusion} &  2023.11  & $1024^2$ & Latent Diffusion (DiT)\\
        &Realistic Vision V5.1 ~\cite{realistivision}& 2023.12 & $512^2$ & Latent Diffusion\\
        &Stable Cascade~\cite{stablecascade} & 2024.02 & $1024^2$ & Latent Diffusion\\
        &Playground V2.5~\cite{li2024playgroundv25insightsenhancing}   & 2024.02  & $1024^2$ & Latent Diffusion  \\
        &ProtoVision V6.6 ~\cite{protovision}  & 2024.03 & $1024^2$ & Latent Diffusion\\
        &Hunyuan~\cite{li2024hunyuanditpowerfulmultiresolutiondiffusion} & 2024.05 &$1024^2$ & Latent Diffusion (DiT)\\
        &SD3~\cite{sd3} & 2024.07 & $1024^2$ & Latent Diffusion (DiT)\\
        &Kolors~\cite{kolors}& 2024.07 & $1024^2$ & Latent Diffusion  \\ 
        &Flux-dev~\cite{flux} & 2024.08& $1024^2$ & Latent Diffusion (DiT) \\ 
        \hdashline
        \multirow{6}{1.7cm}{\centering \textit{Face \\Customization}}& 
        ELITE~\cite{wei2023eliteencodingvisualconcepts}&2023.02& $512^2$ &Latent Diffusion \\
        &FastComposer~\cite{xiao2023fastcomposertuningfreemultisubjectimage} & 2023.05 & $512^2$ & Latent Diffusion\\
        &IP-Adapter-FaceID~\cite{ipa1}& 2023.12 &  $512^2$  & Latent Diffusion \\
        &InstantID~\cite{wang2024instantid}& 2023.12 &  $512^2$  &  Latent Diffusion \\
        &IP-Adapter-FaceID-Plus~\cite{ipa2}  &2023.12  &  $512^2$  &  Latent Diffusion\\
        &PhotoMaker~\cite{li2023photomaker}& 2023.12 &  $512^2$  &  Latent Diffusion  \\
        \hdashline
        \multirow{9}{1.7cm}{\centering \textit{Face\\ Restoration}}&
        
        SPARNet~\cite{chen2020learning}& 2020.12 &  $512^2$&  GAN  \\
        & GPEN~\cite{yang2021gan} & 2021.05 & $512^2$&    GAN\\
        &GFPGAN~\cite{wang2021gfpgan} & 2021.06 & $512^2$ & GAN \\
        &CodeFormer~\cite{zhou2022towards} &2022.08  & $512^2$  & VQ\\
        &VQFR~\cite{gu2022vqfr} & 2022.07 &$512^2$  &  VQ  \\
        &DifFace~\cite{yue2022difface} & 2022.12 &  $512^2$&  Pixel Diffusion \\
        &DR2~\cite{wang2023dr2} & 2023.05 & $512^2$ &  Pixel Diffusion \\
        &DiffBIR~\cite{lin2023diffbir}& 2023.08 &  $512^2$&   Latent Diffusion \\
        &StableSR~\cite{wang2024exploitingdiffusionpriorrealworld} & 2024.06 &  $512^2$  & Latent Diffusion \\
        \bottomrule[1pt]
    \end{tabular}}
\end{table}

\paragraph{Prompt Examples.}
\begin{table}[ht]
    \centering
    \vspace{-5pt}
    \renewcommand{\arraystretch}{1.1} 
    \setlength{\tabcolsep}{4pt}       
    \small                             
    \caption{\textbf{Several examples of prompts for nine categories.}}
    \label{tab:evaluate_prompts}
    \begin{tabular}{m{1.5cm} p{6cm}} 
        \toprule[1pt]
        \textbf{\textit{Category}} & \textbf{Prompt Examples} \\
        \midrule
        \multirow{2}{*}{\centering \textit{General}}  
            & a photo of a woman \\
            & a photo of a middle-eastern man \\
        \hdashline
        \multirow{2}{*}{\centering \textit{Clothing}}   
            & a woman wearing a purple wizard outfit \\
            & a man wearing a hoodie with green stripes \\
        \hdashline
        \multirow{2}{*}{\centering \textit{Accessory}} 
            & a man with black hair styled in a top bun \\
            & an old woman with a vintage hairpin \\
        \hdashline
        \multirow{2}{*}{\centering \textit{Action}} 
            & a woman coding in front of a computer \\
            & a man playing the violin \\
        \hdashline
        \multirow{2}{*}{\centering \textit{Expression}} 
            & a man crying disappointedly, with tears flowing \\
            & a woman looking shocked, mouth wide open \\
        \hdashline
        \multirow{3}{*}{\centering \textit{Background}} 
            & a woman laughing on the lawn \\
            & a young woman with a colorful umbrella stands near a crowd \\
        \hdashline
        \multirow{3}{*}{\centering \textit{View}} 
            & a man wearing a doctoral cap, upper body, with the left side of the face facing the camera \\
            & a man playing the guitar in the view of left side \\
        \hdashline
        \multirow{4}{*}{\centering \textit{Style}} 
            & instagram photo, portrait photo of a man, perfect face, natural skin, hard shadows, film grain \\
            & editorial portrait of a man posing dramatically, sharp lighting, fashion magazine style \\
        \hdashline
        \multirow{4}{*}{\makecell[l]{\textit{Facial} \\ \textit{Attributes}}} 
            & a young girl with large round blue eyes, a flat nose bridge, and purple lipstick \\
            & A man with narrow black eyes, a high nose bridge, a thick beard, and fair skin \\
        \bottomrule[1pt]
    \end{tabular}
\end{table}

The face-centric prompts used for face generation can be categorized into nine classes. Tab.~\ref{tab:evaluate_prompts} presents two example prompts for each category due to space constraints. We ensured equal numbers of prompts for male and female subjects. The prompts for face customization are a subset of those used for face generation.
\noindent \textbf{Degradation Scheme.}
We construct two synthetic degradation pipelines to mimic real-world degradation. The first is first order pipeline following previous works~\cite{chen2020learning,wang2021towards,wang2023dr2}  which can be expressed as 
\begin{equation}
   I_d = [(I \circledast k_\sigma) \downarrow_r + n_\delta]_{{\text{JPEG}}_q}
\end{equation}
High-quality images are degraded through a series of operations, including blurring, downsampling, additive Gaussian noise, and JPEG compression, with respective probabilities of 70\%, 100\%, 20\%, and 70\%. The blur kernel is randomly selected from Gaussian, Average, Median, and Motion blur. The interpolation method is randomly selected from Nearest, Linear, Area, and Cubic interpolation. The downsampling scale factor is randomly chosen from 4, 8, or 16.
The second is a second-order degradation pipeline from previous work~\cite{wang2021realesrgan}. 
\begin{equation}
x = \mathcal{D}^n(y) = (\mathcal{D}_n \circ \cdots \circ \mathcal{D}_2 \circ \mathcal{D}_1)(y).
\end{equation}
Blur, resizing, noise, and JPEG compression are conducted in several orders, along with a sinc filter to simulate common ringing and overshoot artifacts. We used these two pipelines to generate 50\% and 50\% of the synthetic low-quality images, respectively.

\subsection{Additional Examples of \textbf{\textit{FaceQ}} Database}
\begin{figure*}[htbp] 
    \centering
    \vspace{-25pt}
    \includegraphics[width=0.85\textwidth]{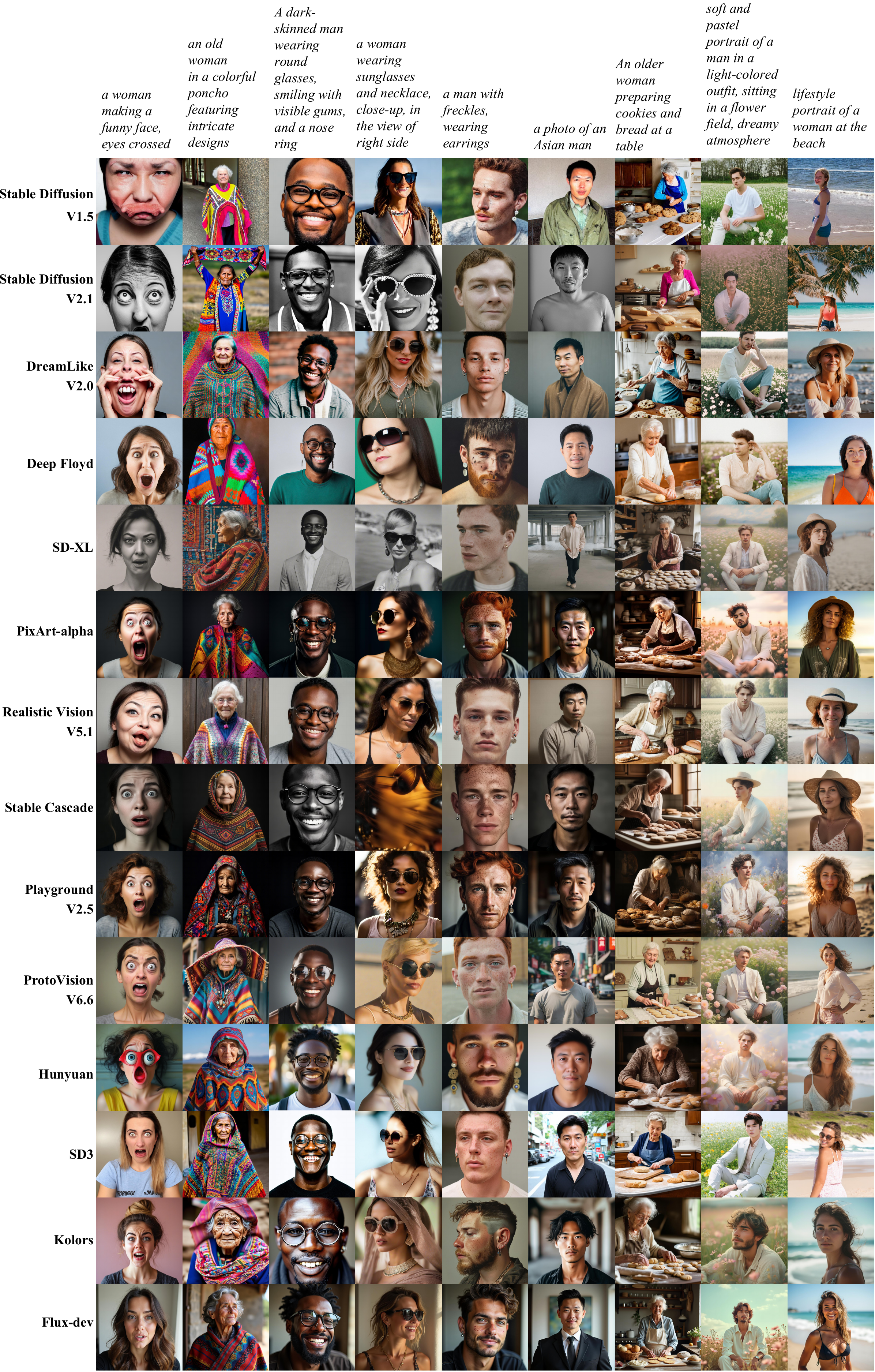} 
    \caption{\textbf{\textit{FaceQ-Gen} Examples.}}  
    \label{fig:gen-ex}  
\end{figure*}

\begin{figure*}
    \centering
    \vspace{-25pt}
    \includegraphics[width=0.85\textwidth]{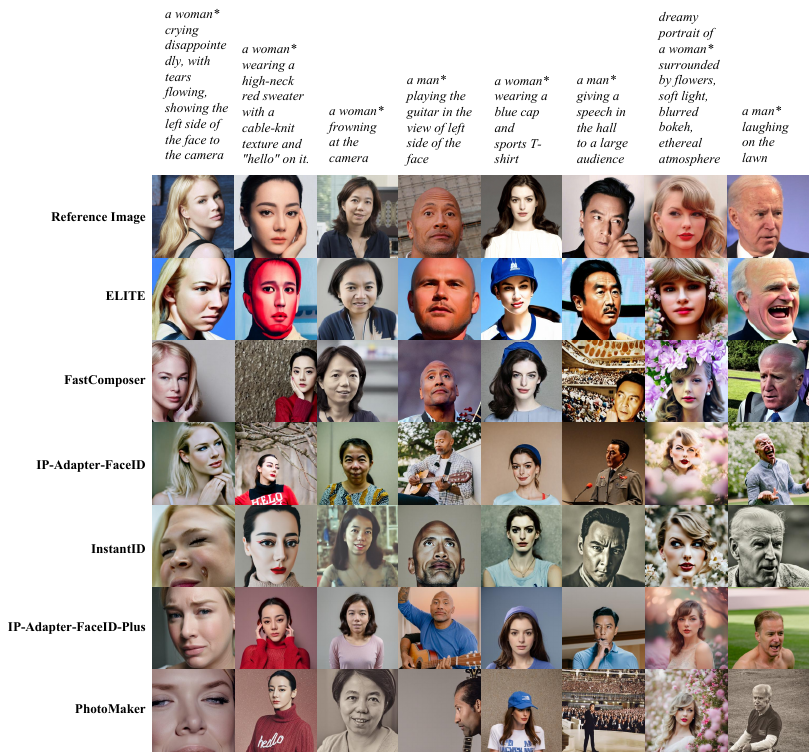}
    \caption{\textbf{\textit{FaceQ-Cus} Examples.}} 
    \label{fig:editex}
\end{figure*}

\begin{figure*}
    \centering
    \includegraphics[width=0.85\textwidth]{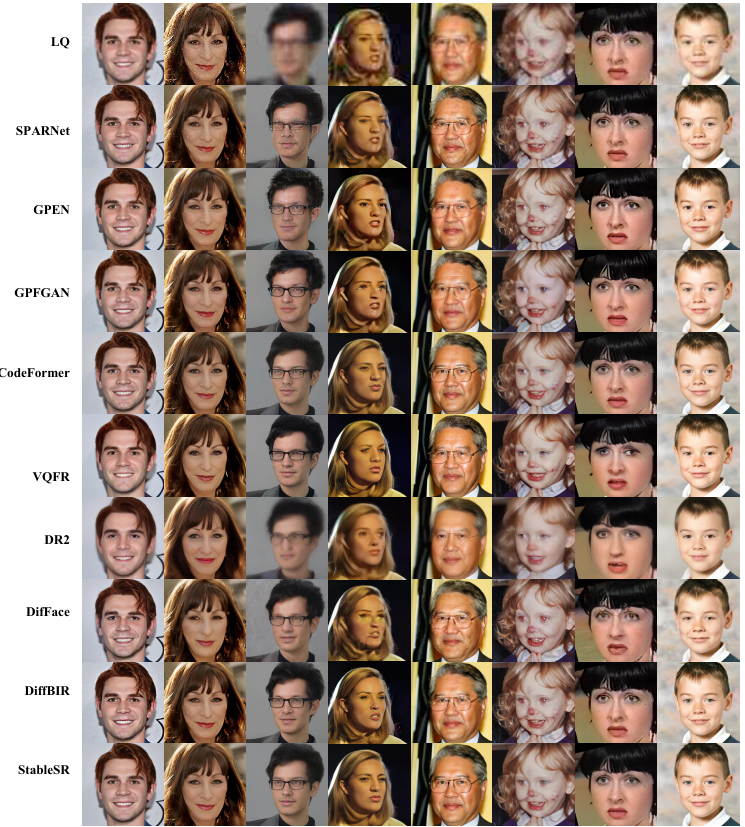}    
    \caption{\textbf{\textit{FaceQ-Res} Examples.}} 
    \label{fig:resex}
   
\end{figure*}

Figure~\ref{fig:gen-ex}, Figure~\ref{fig:editex}, and Figure~\ref{fig:resex} present additional examples from the \textit{FaceQ-Gen}, \textit{FaceQ-Cus}, and \textit{FaceQ-Res} subsets, respectively. Each row corresponds to a specific generative model, showcasing the extensive diversity of content covered by the FaceQ dataset.
\subsection{Quantitative Analysis of \textit{\textbf{FaceQ}} Database}
\begin{figure*}
    \centering
    \includegraphics[width=1\linewidth]{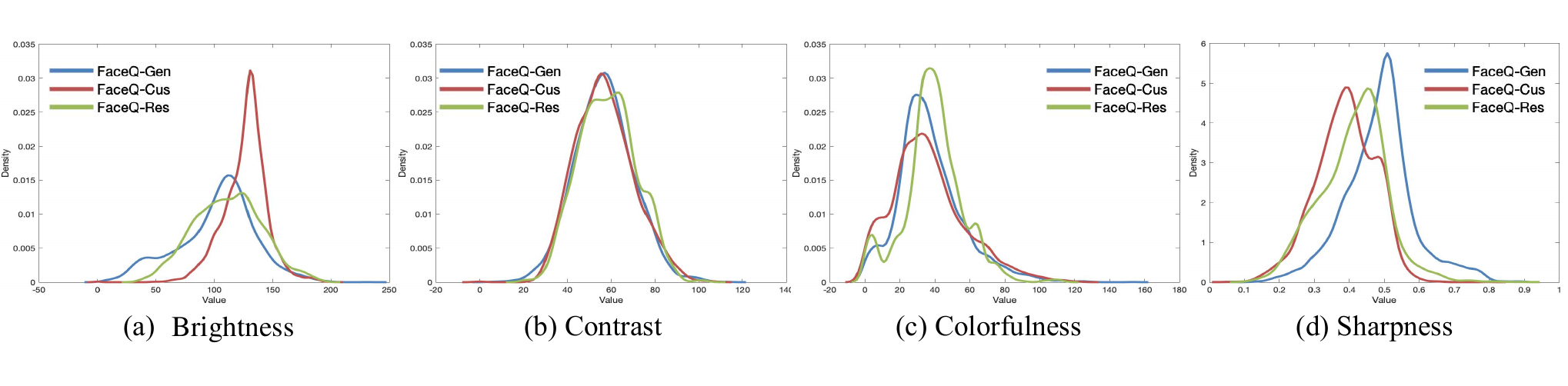}
    \vspace{-25pt}
    \caption{\textbf{Comparisons of the selected four low-level feature distributions calculated on proposed \textit{FaceQ} dataset.}}
    \label{fig:lowlevel}
\end{figure*}

We selected four low-level features—brightness, contrast, colorfulness, and sharpness—to quantitatively assess the content diversity of the \textit{FaceQ} database. Fig.~\ref{fig:lowlevel} illustrates the kernel distribution curves for each selected feature across the three subsets. The results indicate that the images in each subset exhibit a wide range of contrast, colorfulness, and sharpness. However, the \textit{FaceQ-Cus} subset demonstrates a narrower distribution in terms of brightness compared to the other subsets.
We further calculate the relative range $R_i^k$ and coverage uniformity $U_i^k$ of the three subsets across these selected features. The relative range $R_i^k$ is defined as:
\begin{equation}
R_i^k = \frac{{\max (C_i^k) - \min (C_i^k)}}{{{{\max }_k}(C_i^k)}},
\end{equation}
where $C_i^k$ denotes the distribution of $k_\text{th}$ dataset on $i_\text{th}$ feature. ${{\max }_k}(C_i^k)$ refers to the maximum value of $i_\text{th}$ feature across all datasets. The coverage uniformity $U_i^k$ is calculated as the entropy of the B-bin histogram of $C_i^k$ for each subset, using the following formula:  
\begin{equation}
U_i^k =  - \sum\limits_{b = 1}^B {{p_b}{{\log }_B}{p_b},}
\end{equation}
where $p_b$ denotes the normalized number in bin $b$ at  $i_\text{th}$ feature for $k_\text{th}$ dataset. 
\begin{figure*}
    \centering
    \includegraphics[width=1\linewidth]{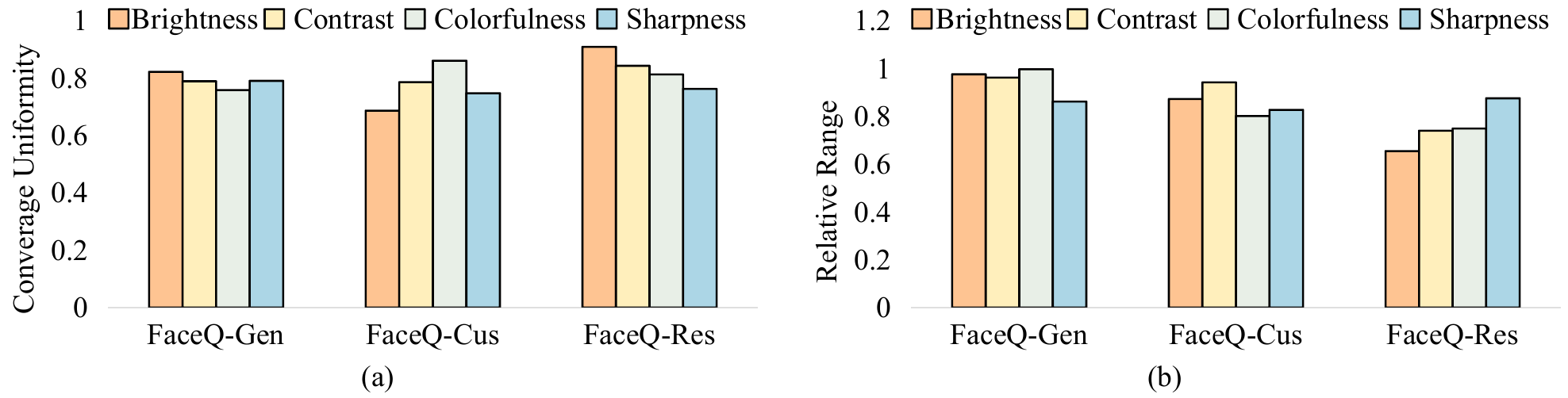}
    \caption{\textbf{Comparisons of the selected four low-level features calculated on the proposed \textit{FaceQ} dataset.} (a) Coverage uniformity. (b) Relative range.}
    \label{fig:ur}
\end{figure*}
Fig.~\ref{fig:ur} presents a quantitative comparison of uniformity and relative range. A higher coverage uniformity indicates a more uniform feature distribution within the database, while a higher relative range reflects greater intra- and inter-dataset differences. It can be observed that all three subsets exhibit a diverse range and a uniform distribution across the four low-level features.
\section{\textit{FaceQ}: Subjective Experiments}
\subsection{Implementation Details}
\begin{figure*} 
    \centering
    \includegraphics[width=1\linewidth]{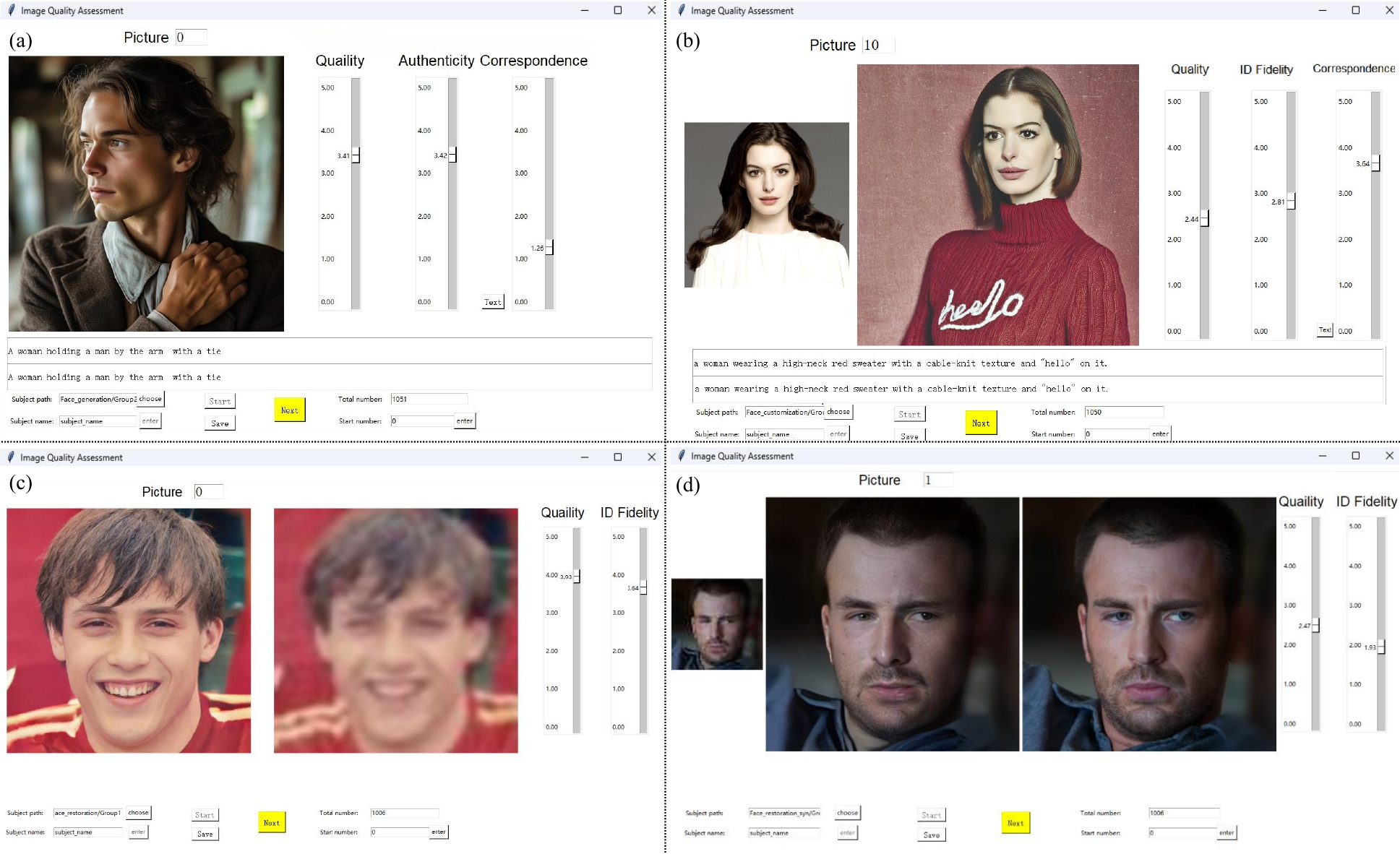}
    \caption{\textbf{Screenshots of the rating interface for human evaluation.} (a) Face generation evaluation interface. (b) Face customization evaluation interface. (c) Face restoration (real world) evaluation interface. (d) Face restoration (synthetic) evaluation interface.}
    \label{fig:ui}
\end{figure*}

Fig.~\ref{fig:ui} presents screenshots of the user rating interfaces for the four tasks. In the generation task, as shown in Figure~\ref{fig:ui} (a), participants are asked to rate images on a scale of 0 to 5 based on \textit{quality}, \textit{authenticity}, and \textit{correspondence}. Prompts are displayed beneath the candidate images, accompanied by translations into the participants' native languages. In the customization task, as shown in Fig.~\ref{fig:ui} (b), the reference image is displayed on the left, with prompts and translation shown below. In Fig.~\ref{fig:ui} (c), the candidate image appears on the left, while the corresponding low-quality reference image is on the right.  In Figure~\ref{fig:ui} (d), both the low-quality image and the ground truth are displayed in synthetic scenarios. Each subset in \textit{FaceQ} was randomly divided into four groups, each containing approximately 1,000 images. Participants were compensated \$14 for completing each group of experiments according to ~\cite{silberman2018responsible}. At last, 3\% invalid data are removed and no subject is removed. 
\subsection{Subjective Evaluation Examples}
\begin{figure*} 
    \centering
    \includegraphics[width=0.95\linewidth]{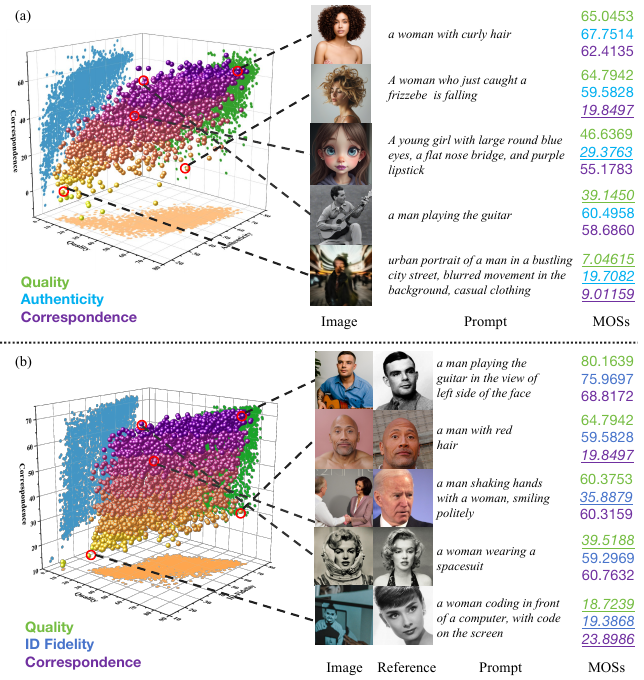}
    \caption{\textbf{Additional visualizations of the 3D scatter of MOSs.} We sample five representative points from the scatter and visualize their MOS scores across three dimensions. Each dimension is represented by a different color. The relatively low dimensions are \underline{\textit{underlined}}. (a) Face generation. (b) Face customization.}
    \label{fig:supp-teaser}
\end{figure*}

Fig.~\ref{fig:supp-teaser} provides a visual supplement to the 3D scatter plots described in the main submission. Fig.~\ref{fig:supp-teaser} (a) presents the 3D scatter plot for the \textit{FaceQ-Gen} subset, showcasing five representative edge points. These images, ranked from top to bottom, correspond to overall good, low correspondence, low authenticity, low quality, and overall bad. As illustrated, the MOS scores effectively and intuitively capture the strengths and weaknesses of the images, accurately reflecting human preferences across different dimensions.
Similarly, Fig.~\ref{fig:supp-teaser} (b) depicts the 3D scatter plot for the \textit{FaceQ-Cus} subset, highlighting another set of five representative points. These images, ranked from top to bottom, correspond to overall good, low correspondence, low identity correspondence, low quality, and overall bad. The MOS score demonstrates a significant decline in dimensions where the image exhibits poor performance. This observation further substantiates the reliability and validity of human scoring in reflecting image quality across multiple dimensions.

\section{\textit{F-Bench}: More Analysis}
\subsection{MOS Distribution}
\begin{figure*}
    \centering
    \includegraphics[width=1\linewidth]{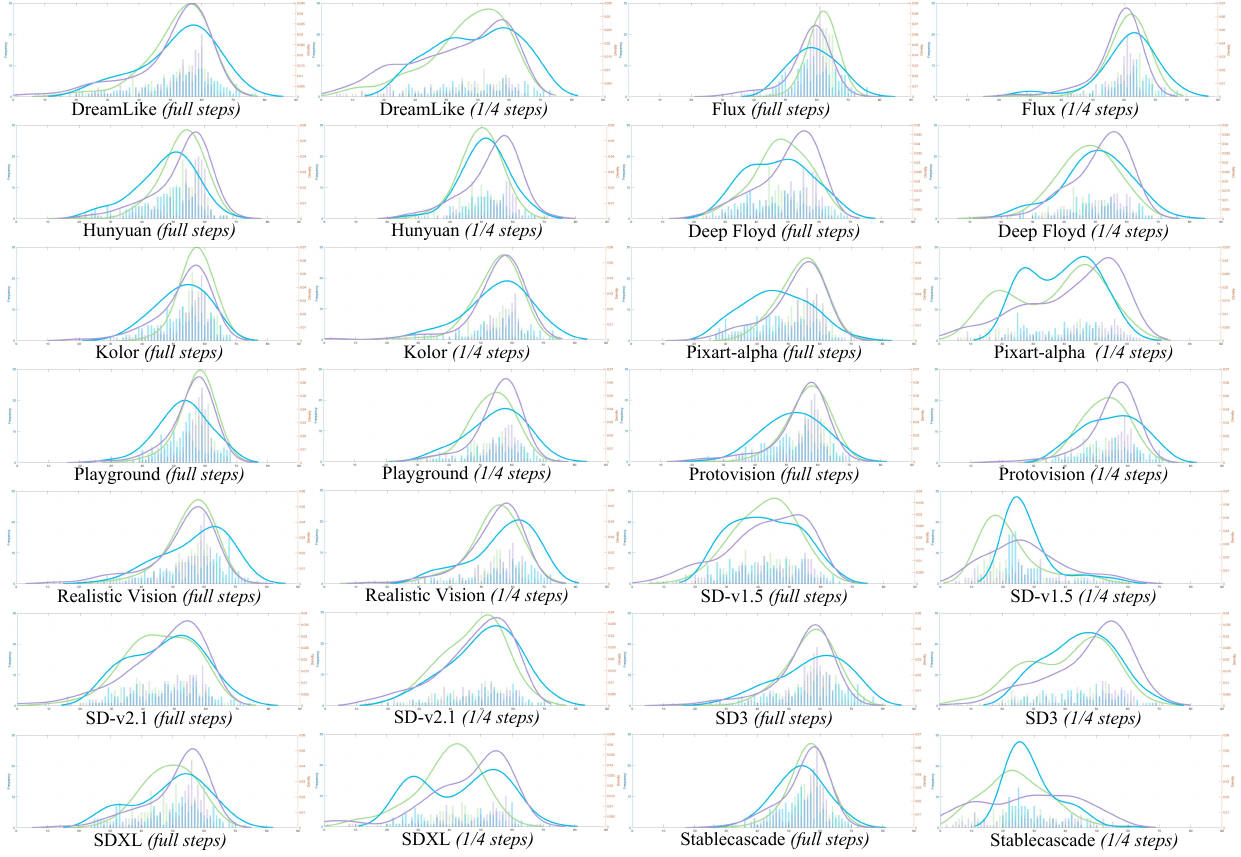}
    \caption{\textbf{MOS distribution histograms and kernel density curves across different \textit{face generation} models.} \textquotedblleft full steps\textquotedblright contains images generated in default sampling steps and \textquotedblleft 1/4 steps\textquotedblright contains the images generated by one-quarter of the default steps.}
    \label{fig:mos1}
\end{figure*}

\begin{figure*}
    \centering
    \includegraphics[width=1\linewidth]{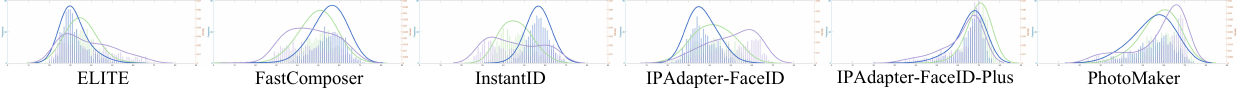}
    \caption{\textbf{MOS distribution histograms and kernel density curves across different \textit{face customization} models.}}
    \label{fig:mos2}
\end{figure*}

\begin{figure*}[h]
    \centering
    \includegraphics[width=1\linewidth]{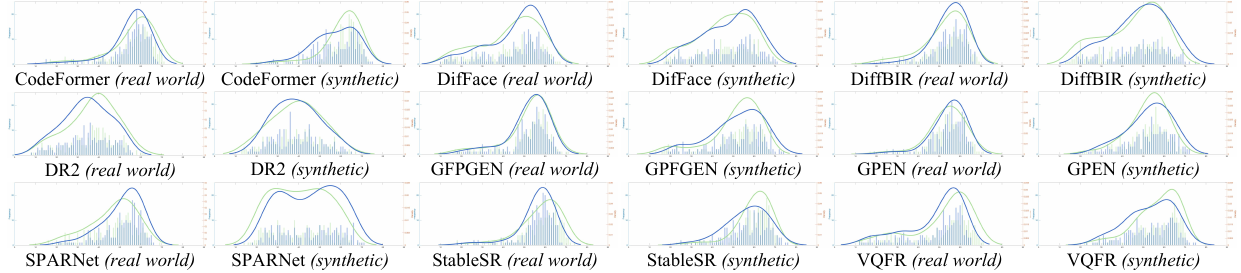}
    \caption{\textbf{MOS distribution histograms and kernel density curves across different \textit{face restoration} models.} \textquotedblleft synthetic\textquotedblright refers to images restored from the synthetic low-quality inputs while \textquotedblleft real world\textquotedblright refers to the images restored from real-world low-quality inputs.}
    \label{fig:mos3}
\end{figure*}

Fig.~\ref{fig:mos1}  illustrates the MOS distributions of all fourteen face generation models across the dimensions of quality, authenticity, and correspondence, for both full-step and 1/4-step performances. The full-step distribution plots provide a comprehensive view of the performance distribution for different methods across the three dimensions, enabling a detailed evaluation of each method's effectiveness. In the 1/4-step distribution plots, it can be observed that models such as Stable Cascade~\cite{stablecascade}, SDXL~\cite{podell2023sdxlimprovinglatentdiffusion}, and Pixart-alpha~\cite{chen2023pixartalphafasttrainingdiffusion} exhibit high sensitivity to the reduction in step size. In contrast, models such as Flux\cite{flux} and RealisticVision\cite{realistivision} demonstrate relatively stable performance with minimal degradation when reducing the steps.
Fig.~\ref{fig:mos2} illustrates the MOS distributions for six face customization models across three dimensions: quality, identity fidelity, and correspondence. Significant variations can be observed among different models and dimensions, highlighting distinct performance characteristics.
Fig.~\ref{fig:mos3} illustrates the MOS distributions for all nine face restoration models across the dimensions of quality and identity fidelity, evaluated for both real-world and synthetic cases. Most models exhibit varying performance between real-world and synthetic inputs, resulting in noticeable differences in their distributions, as exemplified by SPARNet~\cite{chen2020learning}.
\subsection{Perspectiive Analysis}
\begin{figure*}
    \centering
    \vspace{-25pt}
    \includegraphics[width=1\linewidth]{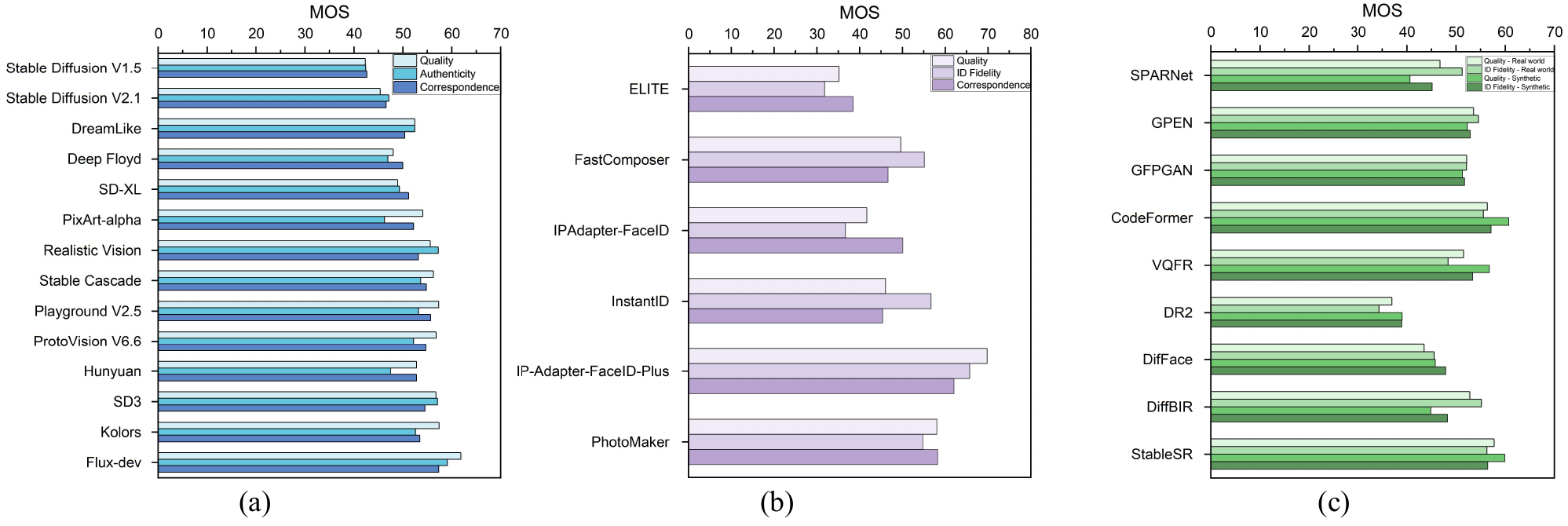}
    \caption{\textbf{Comparison of different model rankings based on the averaged MOS} (a) Face generation models. (b) Face customization models. (c) Face restoration models.}
    \label{fig:rank}
\end{figure*}

To provide a clearer comparison of the strengths and weaknesses of different methods, we present the average MOS scores across various dimensions in Figure~\ref{fig:radar}. For the three dimensions of face generation, \textit{authenticity} exhibits the largest disparity between methods, while \textit{correspondence} and \textit{quality} tend to cluster around higher scores. In the face customization task, the methods show inconsistent performance in \textit{correspondence}, whereas \textit{quality} remains relatively balanced. For face restoration, \textit{quality-synthetic} emerges as the easiest metric to achieve high scores, followed by \textit{quality-real-world}.
Figure~\ref{fig:rank} displays the rankings of the various methods. For face generation, Flux~\cite{flux} achieves the highest performance across all three dimensions. When considering \textit{authenticity}, RealisticVision~\cite{realistivision} and SD3~\cite{sd3} outperform other methods. Playground~\cite{li2024playgroundv25insightsenhancing} ranks second only to Flux~\cite{flux} in terms of \textit{correspondence}, while Kolors~\cite{kolors} and SD3~\cite{sd3} follow Flux~\cite{flux} in \textit{quality}. On the other hand, SDv1.5~\cite{rombach2022high} performs the worst across all dimensions. For face customization, IP-Adapter-FaceID-Plus~\cite{ipa2}, InstantID~\cite{wang2024instantid}, and PhotoMaker~\cite{shan2014photo} excel at preserving identity information. For face restoration, CodeFormer~\cite{zhou2022towards} demonstrates the best performance in synthetic scenarios, while StableSR achieves the highest scores in real-world scenarios.
\begin{figure*}
    \centering
    \includegraphics[width=1\linewidth]{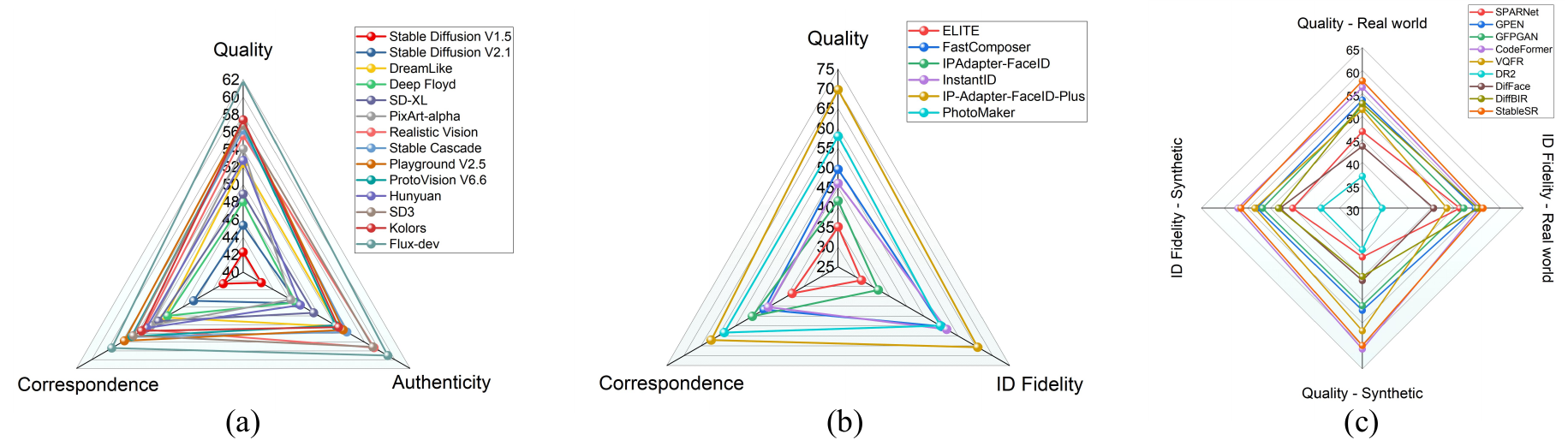}
    \caption{\textbf{Comparison of averaged MOS of different models across \textit{Quality},\textit{Authenticity}, \textit{ID Fidelity}, and \textit{Correspondence}.} (a) Face generation models. (b) Face customization models. (c) Face restoration models.}
    \label{fig:radar}
\end{figure*}

\subsection{Class-wise Comparison}
\begin{figure*}
    \centering
    \vspace{-25pt}
    \includegraphics[width=1\linewidth]{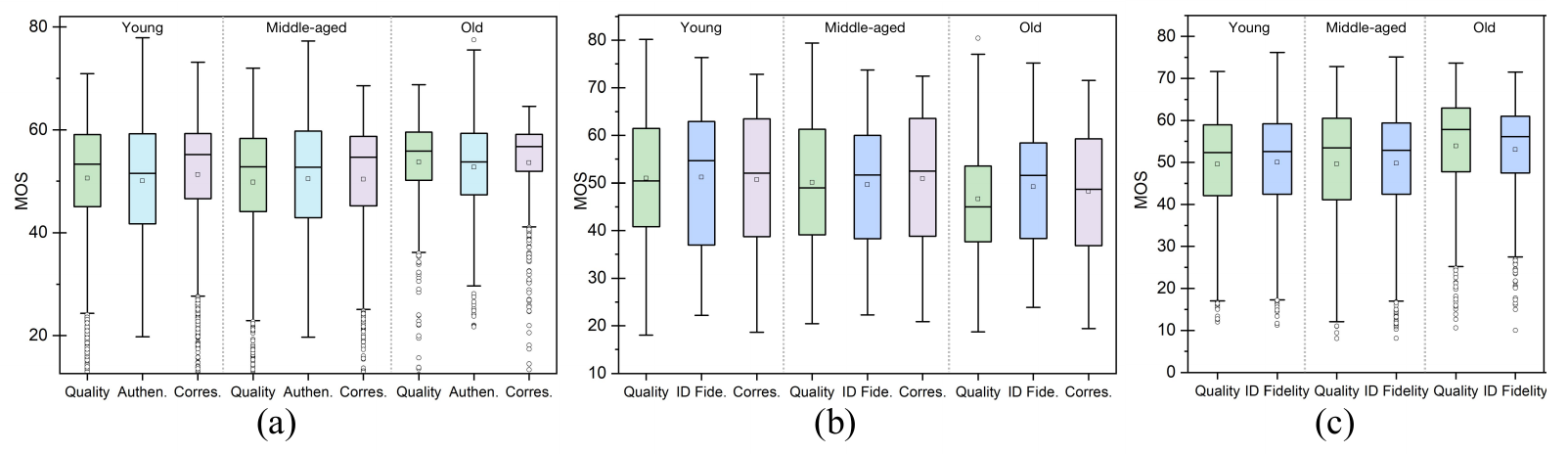}
    \caption{\textbf{Comparison of multi-dimensional MOS distributions across age groups.} \textquotedblleft Authen.\textquotedblright, \textquotedblleft Corres.\textquotedblright and \textquotedblleft ID Fide.\textquotedblright denote \textit{Authenticity}, \textit{Correspondence}, and \textit{ID Fidelity} respectively. (a) Face generation models. (b) Face customization models. (c) Face restoration models.}
    \label{fig:age}
\end{figure*}

\paragraph{Age.} Figure~\ref{fig:age} presents the multi-dimensional MOS distributions across three age groups (Young, Middle-aged, and Old) for face generation, face customization, and face restoration tasks. In the face generation task, the performance across age groups is relatively consistent across all dimensions. For face customization, more pronounced differences are observed, particularly in the \textit{quality} scores, where older individuals exhibit larger variability. In the face restoration task, \textit{quality} scores for old individuals are notably higher compared to middle-aged and young groups, while \textit{identity fidelity} remains relatively consistent. These results highlight that face generation models are less sensitive to age-related factors, whereas face customization and restoration models demonstrate noticeable performance disparities among age groups, especially in dimensions such as ID Fidelity and Quality. The age and gender of the images are labeled  InsightFace. 

\begin{figure*}
    \centering
    \vspace{-25pt}
    \includegraphics[width=1\linewidth]{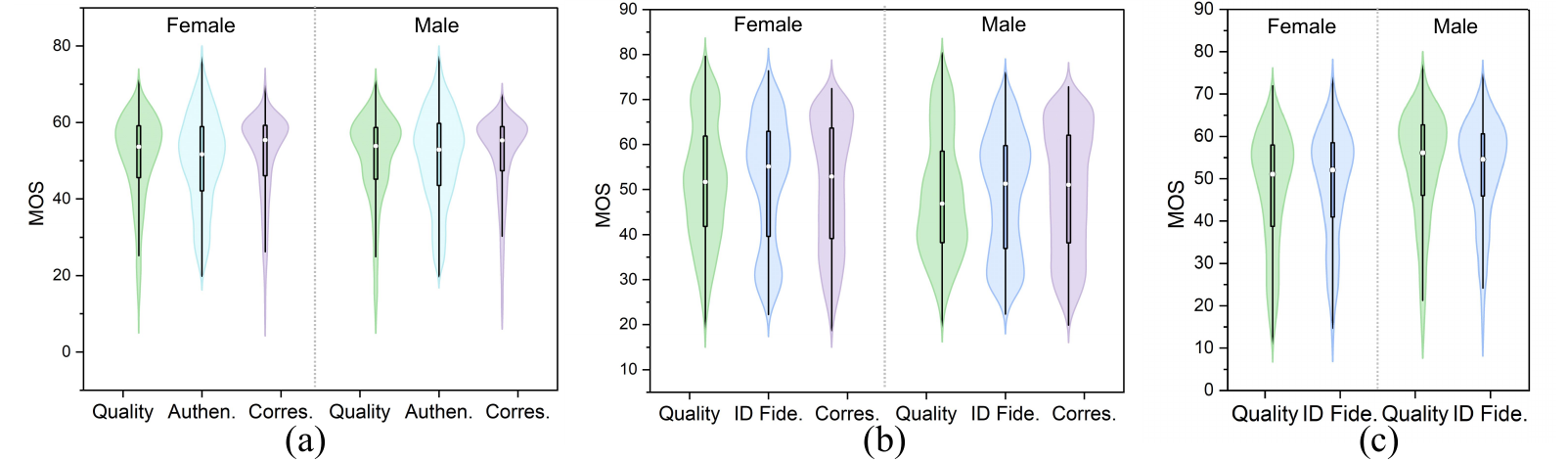}
    \caption{\textbf{Comparison of multi-dimensional MOS distributions across genders.} \textquotedblleft Authen.\textquotedblright, \textquotedblleft Corres.\textquotedblright and \textquotedblleft ID Fide.\textquotedblright denote \textit{Authenticity}, \textit{Correspondence}, and \textit{ID Fidelity} respectively. (a) Face generation models. (b) Face customization models. (c) Face restoration models.}
    \label{fig:gender}
\end{figure*}

\paragraph{Gender.} We visualize the distribution of MOS scores in three dimensions for men and women in each dimension in Fig.~\ref{fig:gender}.
It can be found that the male and female categories in face generation and face customization perform consistently across all evaluation dimensions, with minimal variability observed. However, when it comes to face restoration tasks, the \textit{quality} and \textit{identity fidelity} of the male class are better. This suggests that generation and customization models trained on extensive datasets exhibit less gender bias than restoration models trained on smaller datasets.
\section{QA Methods Implementation Details}

\subsection{Evaluation Metrics}
We adopt three widely used metrics in IQA~\cite{aigciqa,wang2023aigciqa2023largescaleimagequality}: Spearman rank-order correlation coefficient (SRCC), Pearson linear correlation coefficient (PLCC), and Kendall rank correlation coefficient (KLCC) to evaluate the performance of quality assessment methods.

\noindent\textbf{SRCC.}, which ranges from -1 to 1, evaluates the monotonic relationship between two variables. For \( N \) images, it is computed as:

\begin{equation}
SRCC = 1 - \frac{{6\sum\nolimits_{n = 1}^N {{{({v_n} - {p_n})}^2}} }}{{N({N^2} - 1)}},
\end{equation}

Here, \( v_n \) represents the rank of the ground truth value \( y_n \), while \( p_n \) corresponds to the rank of the predicted value \( \hat{y}_n \). When the SRCC value is higher, it signifies a stronger monotonic agreement between the ground truth and the predicted scores. \\
\noindent\textbf{PLCC.} quantifies the linear correlation between predicted scores and ground truth scores and is formulated as:
\begin{equation}
PLCC = \frac{{\sum\nolimits_{n = 1}^N {({y_n} - \bar{y})({{\hat{y}}_n} - \bar{\hat{y}})} }}{{\sqrt {\sum\nolimits_{n = 1}^N {{{({y_n} - \bar{y})}^2}} } \sqrt {\sum\nolimits_{n = 1}^N {{{({{\hat{y}}_n} - \bar{\hat{y}})}^2}} } }},
\end{equation}
where \( \bar{y} \) and \( \bar{\hat{y}} \) denote the mean values of the ground truth scores and the predicted scores, respectively. \\
\noindent \textbf{KLCC.} measures the ordinal association between two measured quantities and is defined as:
\begin{equation}
KLCC = \frac{{2(C - D)}}{{N(N-1)}},
\end{equation}
where \( C \) is the number of concordant pairs and \( D \) is the number of discordant pairs among all possible pairs of \( N \) data points. A higher KLCC indicates a stronger rank correlation between the two variables. Together, these metrics provide a comprehensive evaluation of the relationship between predicted preference scores and ground truth MOS values across different aspects of correlation.
\section{More Details of Our F-Eval Model}
\subsection{Loss Funtion}
We use both language loss, L1 loss and cross-entropy loss as the loss functions to optimize the training process. Specifically, the language loss is used to restrict the F-Eval to produce specific quality-related answer patterns. The language loss function can be formulated as:
\begin{equation}
    \mathcal{L}_{\text{language}} = -\frac{1}{N} \sum_{i=1}^{N} \log P(y_{\text{label}} | y_{\text{pred}})
\end{equation}

where \( y_{\text{pred}} \) is the predicted token, \( y_{\text{label}} \) is the ground truth token, \( P(y_{\text{label}} | y_{\text{pred}}) \) indicates the probability, and \( N \) is the number of tokens.\\
 L1 loss is used to regress the quality scores. The L1 loss can be formulated as:
\begin{equation}
\mathcal{L}_1 = \frac{1}{N} \left| q_{\text{pred}} - q_{\text{label}} \right|,
\end{equation}
where \( q_{\text{pred}} \) is the predicted quality score, \( q_{\text{label}} \) is the ground truth quality score, and \( N \) is the number of images in a batch. \\
Cross-entropy loss to predict the dimension ID  from the input text tokens. The cross-entropy loss can be formulated as:
\begin{equation}
\mathcal{L}_{\text{CE}} = -\frac{1}{N} \sum_{i=1}^{N} \sum_{c=1}^{C} y_{i,c} \log(p_{i,c}),
\end{equation}
where \( y_{i,c} \) is the dimension label for the \( i \)-th sample in class \( c \) (dimension ID), \( p_{i,c} \) is the predicted probability for the \( i \)-th sample in class \( c \), and \( N \) \( N \) is the number of images in a batch. \( C \) is 4.
The overall loss function can be formulated as:
\begin{equation}
\mathcal{L} = \mathcal{L}_{\text{language}} + \mathcal{L}_1+\mathcal{L}_{\text{CE}},
\end{equation}
\begin{table*}[!t]
\centering
\huge
\caption{Complete results in the ablation study.}
\vspace{-3pt}
\label{tab:suptab}
\renewcommand{\arraystretch}{1.1} 
\resizebox{\textwidth}{!}{ 

\begin{tabular}{l|| c c c: c c c: c c c|| c c c: c c c }
\toprule[2pt]
\rowcolor{gray!20} 
Task &\multicolumn{3}{c:}{}&\multicolumn{3}{c:}{\textbf{Face Generation}} &\multicolumn{3}{c||}{}&\multicolumn{6}{c}{\textbf{Face Restoration syn}}\\
\midrule
Dimension & \multicolumn{3}{c:}{Quality} & \multicolumn{3}{c:}{Authenticity} & \multicolumn{3}{c||}{Correspondence}& \multicolumn{3}{c:}{Quality} & \multicolumn{3}{c}{ID Fidelity} \\
\cmidrule{2-4} \cmidrule{5-7} \cmidrule{8-10}\cmidrule{11-13}\cmidrule{14-16}
Method & SRCC$\uparrow$ & KRCC$\uparrow$ & PLCC$\uparrow$ & SRCC$\uparrow$ & KRCC$\uparrow$ & PLCC$\uparrow$ & SRCC$\uparrow$ & KRCC$\uparrow$ & PLCC$\uparrow$ & SRCC$\uparrow$ & KRCC$\uparrow$ & PLCC$\uparrow$ & SRCC$\uparrow$ & KRCC$\uparrow$ & PLCC$\uparrow$  \\
\midrule
Freeze Projector & 0.8080 & 0.6428 & 0.8698 & 0.7518 & 0.5832 & 0.8123 & 0.8475 & 0.6783 & 0.9031 &0.7769&0.5787&0.8074&0.7757&0.5832&0.8074 \\
Freeze Face encoder & 0.8151 & 0.6450 & 0.8733 & 0.7524 & 0.5834 & 0.8140 & 0.8347 & 0.6654 & 0.8912 &0.8121& 0.6120&0.8430 &0.7828& 0.6021& 0.8341\\
Single LoRA & 0.7927 & 0.6140 & 0.8659 & 0.7757 & 0.5995 & 0.8285 & 0.8191 & 0.6429 & 0.8760&0.8377&0.6538&0.8874&0.8228&0.6341&0.8556 \\
Task LoRA & 0.8231 & 0.6477 & 0.8895 & 0.7803 & 0.6056 & 0.8317 & 0.8377 & 0.6661 & 0.8943 &0.8244&0.6428&0.8806&0.8257&0.6359&0.8519\\
F-Eval (Ours) & 0.8486 & 0.6670 & 0.9085 & 0.8312 & 0.6585 & 0.8578 & 0.8471 & 0.6637 & 0.9106 &0.8692&0.6855&0.9009&0.8507&0.6731&0.8726\\
\midrule 
\rowcolor{gray!20} 
Task &\multicolumn{9}{c||}{\textbf{Face Customization}}&\multicolumn{6}{c}{\textbf{Face Restoration rw}}\\
\midrule
Dimension & \multicolumn{3}{c:}{Quality} & \multicolumn{3}{c:}{Authenticity} & \multicolumn{3}{c||}{Correspondence}& \multicolumn{3}{c:}{Quality} & \multicolumn{3}{c}{ID Fidelity} \\
\cmidrule{2-4} \cmidrule{5-7} \cmidrule{8-10}\cmidrule{11-13}\cmidrule{14-16}
Method & SRCC$\uparrow$ & KRCC$\uparrow$ & PLCC$\uparrow$ & SRCC$\uparrow$ & KRCC$\uparrow$ & PLCC$\uparrow$ & SRCC$\uparrow$ & KRCC$\uparrow$ & PLCC$\uparrow$ & SRCC$\uparrow$ & KRCC$\uparrow$ & PLCC$\uparrow$ & SRCC$\uparrow$ & KRCC$\uparrow$ & PLCC$\uparrow$ \\
\midrule
Freeze Projector & 0.9242 & 0.7651 & 0.9233 & 0.9269 & 0.7679 & 0.9242 & 0.8429 & 0.6726 & 0.8386&0.7067&0.5249&0.7114&0.4778&0.3616&0.5451 \\
Freeze Face encoder & 0.9239 & 0.7662 & 0.9217 & 0.9214 & 0.7612 & 0.9208 & 0.8550 & 0.6836 & 0.8524&0.7158& 0.5421& 0.8532& 0.6028& 0.4386& 0.6898\\
Single LoRA & 0.9419 & 0.7938 & 0.9421 & 0.9424 & 0.7957 & 0.9430 & 0.8850 & 0.7140 & 0.8974&0.8019&0.6155&0.8455&0.6710&0.5017 & 0.7768\\
Task LoRA & 0.9421 & 0.7984 & 0.9403 & 0.9422 & 0.7985 & 0.9402 & 0.8727 & 0.7005 & 0.8918 &0.7540&0.5661&0.7964&0.6376&0.4736&0.7403\\
F-Eval (Ours) & 0.9462 & 0.7961 & 0.9461 & 0.9188 & 0.7640 & 0.9322 & 0.9460 & 0.7959 & 0.9457 &0.8448&0.6577&0.8705&0.7957&0.6057&0.8366\\

\bottomrule[2pt]
\end{tabular}}
\vspace{-8pt}
\end{table*}
\subsection{Failure Cases} F-Eval’s performance may degrade when the face region in the image is either too small or in an extreme side view. In these scenarios, the face features extracted by the face encoder become less accurate, leading to a drop in performance, particularly in the quality and authenticity dimensions. We will address this limitation by integrating a more robust face encoder fine-tuned on side view and small face data in future works. Additionally, F-Eval currently does not support identifying specific distorted regions, which will also be addressed in future works.

\subsection{More Experimental Results}
The detailed ablation study results are listed in Tab.~\ref{tab:suptab}. The results indicate the effectiveness of the key components in our F-Eval.

\label{sec:sup-dataset}

\end{document}